\documentclass{article} %
\usepackage{iclr2027_conference,times}
\usepackage{amsmath,amsfonts,bm}

\def\eqref#1{equation~\ref{#1}}
\def\1{\bm{1}}

\DeclareMathAlphabet{\mathsfit}{\encodingdefault}{\sfdefault}{m}{sl}
\SetMathAlphabet{\mathsfit}{bold}{\encodingdefault}{\sfdefault}{bx}{n}

\usepackage{xcolor}
\usepackage[normalem]{ulem}
\newcommand{\fixnew}[1]{#1}
\newcommand{\fixcut}[1]{}
\definecolor{todocol}{RGB}{181,125,0}

\usepackage{hyperref}
\definecolor{linkcol}{RGB}{26,64,140}
\hypersetup{colorlinks=true, linkcolor=linkcol, citecolor=linkcol,
            urlcolor=linkcol, pdfborder={0 0 0}}
\usepackage{url}
\usepackage{booktabs}
\usepackage{multirow}
\usepackage{tikz}
\usetikzlibrary{arrows.meta,calc}
\usepackage{amsmath,amsthm}
\newtheorem{proposition}{Proposition}

\newtheorem{lemma}[proposition]{Lemma}
\theoremstyle{remark}\newtheorem{remark}{Remark}

\title{Keep the Lineages Alive: Diversity-Preserving SMC for Diffusion LLM Decoding}

\author{\textbf{Minoo Ahmadi} \quad \textbf{Seyedarmin Azizi} \quad \textbf{Mehdi Kamal} \quad \textbf{Massoud Pedram} \\
University of Southern California, Los Angeles, USA \\
\texttt{\{minooahm, seyedarm, mehdi.kamal, pedram\}@usc.edu}}

\iclrfinalcopy
\begin{document}
\maketitle
\lhead{Preprint}

\begin{abstract}
Inference-time sequential Monte Carlo (SMC) decoding can improve the reasoning accuracy of
masked diffusion language models (diffusion LLMs) without post-training by resampling a population
of partial generations according to a reward. Its effectiveness therefore depends on the resampling
step, which must concentrate computation on promising candidates while retaining enough diversity
to explore alternative reasoning paths. The reward can be internal, the decoder's own confidence on
the committed prefix, or external, a process reward model scoring that prefix. Under either reward,
multinomial resampling can disrupt this balance by duplicating high-weight particles and discarding
lineages that may still reach a correct solution. The effective sample size (ESS) does not register
this loss: multinomial resampling resets it to its maximum even as distinct lineages disappear. To
retain such lineages, we introduce lineage-preserving SMC (LP-SMC), which combines Chopthin
resampling with an input weight floor. Instead of equalizing weights, Chopthin bounds the ratio
between the largest and smallest weights it returns and carries the unequal weights forward. The
floor limits how far below the maximum a weight may fall before resampling, so that lineages with
low early weight keep a chance to recover. Across mathematical-reasoning and code-generation
benchmarks on two base models, LP-SMC raises oracle coverage, the fraction of problems with at
least one correct particle, in every setting, by $1.8$ to $25.6$ points, and raises end-to-end
accuracy under a common reward-model selector in eleven of twelve settings, by up to $7.9$ points
on LLaDA-1.5 and $18.3$ on Dream-7B. The coverage gain persists under the external reward model,
which suggests that diversity-preserving resampling and the guidance signal contribute separately
to inference-time search.
\end{abstract}

\input{fig_overview}
% ------------------------------------------------------------------
% Introduction — v8 (user's LaTeX keys and labels; three result
% sentences settled or bracketed, 19 Sep 2026)
% Six paragraphs + four contributions + prior-work note.
% Spine: rewards determine which trajectories appear promising;
% resampling determines which trajectories remain available to become
% correct. No formulas in the introduction.
%
% Brackets open when the runs land:
%   [B] multinomial+PRM on MATH-500 (110/500, ~15 h) -> paragraph 6 and
%       the Evaluation bullet ("separately test whether" -> "show that").
% [B] was rewritten on 19 Sep after the direct measurement from
% unique_roots_post: the earlier "factor of five" composition claim used
% distinct texts as a proxy for lineages and is withdrawn. Measured: the
% resampler multiplies surviving lineages by ~2 (2.6x under confidence,
% 2.0x under the PRM, n=113); the reward multiplies texts-per-lineage by
% ~1.33 under either resampler; Chopthin multiplies texts-per-lineage by
% ~0.6 under either reward (fewer copies per lineage).
% Verified 19 Sep against the logs: 2.58 / 1.96 on lineages, 1.33 / 1.35
% on texts-per-lineage, 0.59 / 0.60 for Chopthin. The two matched pairs
% agree to within 0.02, so "close to independent" is if anything
% understated. Lineages 2.71 -> 6.98 of 16; texts 6.90 -> 10.53.
% ------------------------------------------------------------------

\section{Introduction}
% 2026-09-25: Introduction trimmed for the page budget (approved rewrite with corrections).
% Earlier accepted \fixnew wrappers were unwrapped so that this round's marks are the only
% ones here; struck passages carry plain-text citations and cross-references because
% \sout cannot hold \cite or \ref. The four-bullet contributions list is reproduced as
% running text inside \fixcut for the same reason.

Inference-time scaling can improve language-model reasoning by allocating
more computation during decoding \citep{brown2024monkeys,snell2024scaling}\fixcut{.
Sequential Monte Carlo (SMC) provides a probabilistic framework for this setting (Doucet et
al., 2001), and has been applied to autoregressive language models using learned twist
functions or process rewards (Zhao et al., 2024; Puri et al., 2025), modified target
distributions (Azizi et al., 2026), and diffusion language models (Singhal et al.,
2025).}\fixnew{, and sequential Monte Carlo (SMC) provides a probabilistic framework for it
\citep{doucet2001sequential}, applied to autoregressive language models with learned twist
functions or process rewards \citep{zhao2024twistedsmc,puri2025reasoningsmc} and to
text diffusion models \citep{singhal2025general}.} In particular, self-rewarding SMC
(SR-SMC; \citealp{luo2026srsmc}) weights each particle by the confidence accumulated along
its own trajectory and improves diffusion LLMs on mathematical reasoning and code
generation without further training.

\fixcut{An important source of information loss, however, can arise during resampling
itself. Recent SMC decoding methods have primarily introduced new proposal or weighting
mechanisms while typically retaining standard resampling schemes. SR-SMC uses multinomial
resampling, which converts continuous particle weights into a discrete number of
descendants (Douc et al., 2005), replicating high-weight particles and discarding
low-weight ones. In reasoning tasks, a trajectory that appears less promising at an
intermediate stage may still develop into a correct completion, so eliminating it
according to its current weight can be premature: resampling removes trajectories whose
weight is currently low, not necessarily those that will ultimately be incorrect.
Figure 1 illustrates this effect for a four-particle population, where the highest-weight
lineage (the particles descending from one initial particle) is incorrect while a
lower-weight correct lineage is removed. Declining to resample avoids the loss entirely
but forfeits the concentration of computation that motivates SMC; the question is how to resample
without discarding what the population has already reached.}
\fixnew{Information can be lost during resampling itself. Recent SMC decoding methods
introduce new proposal or weighting mechanisms while retaining standard resampling
schemes; SR-SMC uses multinomial resampling, which converts continuous weights into a
discrete number of descendants \citep{douc2005comparison}, replicating high-weight
particles and discarding low-weight ones. In reasoning tasks a lineage that carries
low intermediate weight may still develop into a correct completion, so
resampling removes trajectories whose weight is currently low, not necessarily those that
will ultimately be incorrect. Figure~\ref{fig:overview} illustrates this for a
four-particle population, where the highest-weight lineage (the particles descending from
one initial particle) is incorrect while a lower-weight correct lineage is removed.
Declining to resample avoids the loss but forfeits the concentration of computation that motivates SMC; the question is how to resample without discarding what the population has
already reached.}

\fixcut{In SMC, which trajectories survive a resampling event is decided from the
trajectory weights by the rule of the particular resampler. Improving the reward changes
the weights; it does not change the rule. A stronger process reward model gives more
informative weights, but multinomial resampling still converts them into a finite set of
descendants and discards the rest. The standard diagnostic, the effective sample size
(ESS), does not measure this loss either. Repeated resampling is known to reduce path
diversity (Kitagawa, 1996; Doucet et al., 2001), yet the ESS measures weight imbalance, not how many lineages remain, and multinomial resampling resets it to its maximum
while lineages disappear. We formalize this mismatch and show that multinomial resampling
of equally weighted particles from distinct lineages eliminates roughly one third of them
in expectation.}
\fixnew{A better reward does not resolve this: improving the reward changes the weights, not the rule that converts them into
descendants: a stronger process reward model still leaves multinomial resampling to
discards the low-weight particles. Nor does the standard diagnostic, the effective sample size (ESS),
measure this loss. Repeated resampling is known to reduce path diversity
\citep{kitagawa1996monte,doucet2001sequential}, yet the ESS measures weight imbalance, not how many lineages remain, and multinomial resampling resets it to its maximum
while lineages disappear. We formalize this mismatch and show that multinomial resampling
of equally weighted particles from distinct lineages eliminates roughly one third of them
in expectation.}

\fixcut{Preserving trajectory diversity requires addressing two distinct failure modes.
First, a moderately weighted lineage may disappear solely because of the stochasticity of
resampling. Second, the weight of a trajectory may fall so far below the maximum before
resampling that it is unlikely to survive even under a more diversity-preserving rule,
which is common in block-wise decoding because the weight is a product over many
committed tokens (Section 2.2). We address the first with Chopthin resampling (Gandy and
Lau, 2016), which bounds the ratio between the returned weights instead of equalizing them
and carries the unequal weights forward. We address the second with a weight floor that
limits how far below the maximum the weight of any particle may fall before resampling.
Neither modification requires additional model evaluations. Together they form
lineage-preserving SMC (LP-SMC). We derive the resulting ESS guarantee, the admissible
range of the floor, and the bias introduced by flooring in Section 4.}
\fixnew{Preserving lineage diversity requires addressing two failure modes. First, a
moderately weighted lineage may disappear solely through the stochasticity of resampling.
Second, a lineage's weight may fall so far below the maximum before resampling that it
is unlikely to survive even under a diversity-preserving rule, which is common in
block-wise decoding because the weight is a product over many committed tokens
(Section~\ref{sec:tilt}). We address the first with Chopthin resampling
\citep{gandy2016chopthin}, which bounds the ratio between the returned weights instead of
equalizing them and carries the unequal weights forward, and the second with a weight
floor that limits how far below the maximum any weight may fall before resampling.
Neither requires additional model evaluations. Together they form lineage-preserving SMC
(LP-SMC); Section~\ref{sec:method} derives the resulting ESS guarantee, the gate condition
on the floor and the bias it introduces.}

\fixcut{We keep the proposal mechanism fixed and evaluate the population-management
strategy under two forms of guidance: the internal self-reward used by SR-SMC, derived
from sequence-level model confidence, and an external process reward model. We also
evaluate several selection rules separately, including majority-based selection and
reward-model ranking, because constructing a useful candidate population and identifying
its best member are different problems.}

\fixcut{Across four mathematical reasoning and code-generation benchmarks, the proposed
population-management strategy consistently increases both genealogical diversity,
measured as surviving founding lineages and distinct outputs, and oracle coverage, the
fraction of problems for which at least one of the $N$ particles is correct. These gains
generally translate into higher final-answer accuracy, although the magnitude depends on
the selection rule. On MATH-500 oracle coverage improves by $9.4$ points and on Dream-7B
HumanEval by $25.6$; on GSM8K, the largest test set, surviving lineages rise from $5.2$ to
$10.0$ (Table 2) and accuracy under the reward-model selector by $2.3$ points (Table 4).
Oracle coverage improves in every setting, on both base models.}
\fixnew{We keep the proposal fixed and evaluate population management under two forms of
guidance, the internal self-reward of SR-SMC and an external process reward model, and
under several selection rules, because constructing a population that contains a correct answer and selecting it are different problems. Across twelve model--guidance--benchmark settings,
LP-SMC increases oracle coverage, the fraction of problems for which at least one of the
$N$ particles is correct, in all twelve. In the internal-guidance ablations, it also
increases surviving lineages and distinct outputs across both models and all reported
benchmarks; on GSM8K with LLaDA-1.5, surviving lineages rise from $5.2$ to $10.0$ of $16$
(Table~\ref{tab:ablation2}). How much of that coverage becomes final accuracy, however, depends on the selection rule.}

\fixcut{Our contributions are as follows. Diagnosis. We formalize a mismatch between the
ESS criterion and lineage diversity in SMC decoding: multinomial resampling can eliminate
distinct lineages while resetting ESS to its maximum (Section 3.1). Method. We introduce
Chopthin resampling into block-wise masked diffusion SMC, carrying the unequal
post-resampling weights forward, together with a complementary weight floor that prevents
a trajectory from falling arbitrarily far below the maximum before resampling. Neither
requires additional model evaluations. We derive the ESS lower bound implied by Chopthin's
weight-ratio bound (Lemma 2), tighten the bound of Gandy and Lau (2016) by a constant, and
characterize the floor's admissible range and the bias it introduces (Sections 4.1 and
4.2). Evaluation. Across four reasoning and code-generation benchmarks and two base models
(LLaDA-1.5 and Dream-7B), we evaluate the method under internal guidance and separately
under external process-reward guidance, using a common reward-model selection rule for the
headline comparison, while analyzing alternative selection rules separately (Section 5).
Population versus selection. We separate how the population is built from how an answer is
selected from it, and show how gains in coverage translate into accuracy under different
selection rules (Section 5.3). Chopthin has been applied to autoregressive power sampling
(Ahmadi et al., 2026); the block-wise diffusion setting, the weight floor, the
ESS-versus-lineage analysis and the selector study are new here.}
\fixnew{\paragraph{Contributions.} (i)~We formalize the mismatch between the ESS criterion
and lineage diversity: multinomial resampling can eliminate distinct lineages while
resetting the ESS to its maximum (Section~\ref{sec:motivation}). (ii)~We address the two complementary sides of a resampling event: the weight floor bounds
how far a lineage may collapse \emph{before} the event, and Chopthin with carried weights
bounds the ratio of the weights returned \emph{by} it. We introduce both into block-wise
masked diffusion SMC, derive the ESS lower bound implied by Chopthin's ratio bound,
tightening that of \citet{gandy2016chopthin} by a constant, and characterize the floor's
gate condition and bias (Section~\ref{sec:method}). (iii)~We evaluate LP-SMC on four
benchmarks and two base models, under internal guidance on all four benchmarks and under
external process-reward guidance on the mathematics benchmarks, with a common reward-model
selector for the headline comparison and a separate analysis of how coverage converts to
accuracy under alternative selection rules (Section~\ref{sec:experiments}).}

\section{Background: particle decoding for masked diffusion \fixcut{LMs}\fixnew{LLMs}}
\label{sec:background}

\subsection{Masked diffusion decoding}
\label{sec:decoding}

Unlike autoregressive decoding, which extends a sequence from left to right, a
diffusion LLM \citep{sahoo2024mdlm, llada2025} starts from a fully masked sequence
and replaces masked positions with tokens over a series of steps, several positions at a
time \fixnew{(Appendix~\ref{app:notation} collects the notation)}.
\paragraph{Setup.}
A sequence has $L$ positions and tokens from a vocabulary $\mathcal{V}$. Decoding runs
backward over $T$ steps from the fully masked $x_T$ to the output $x_0$. The positions are
partitioned into consecutive blocks of fixed size; only positions of the current block can
be committed, \fixcut{and }decoding moves on when the block is complete\fixnew{, and a
particle whose block is complete waits until every particle has completed it. Resampling
is considered at these boundaries}.

\paragraph{One step.}
With $\mathcal{M}_t$ the masked positions of the current block, a step
\fixcut{(i)~}draws $\hat x_0(j) \sim p_\theta(\cdot \mid x_t)$ at every $j \in \mathcal{M}_t$,
independently, at temperature one, where $p_\theta(\cdot \mid x_t)$ is the model's
predictive distribution given the partially masked sequence $x_t$ (the prompt is
implicit)\fixcut{; (ii)~records}\fixnew{. It records} the confidence
$c_t(j) = p_\theta(\hat x_0(j) \mid x_t)$\fixcut{; (iii)~commits}\fixnew{. It commits} the
set $\mathcal{S}_t$ of positions whose confidence exceeds a threshold (the most confident
one if none does; ties by a fixed rule)\fixcut{; (iv)~forms}\fixnew{. It forms} $x_{t-1}$
by writing the drawn tokens at $\mathcal{S}_t$ and copying everything else. Committed
tokens are never revised.

\fixcut{\textbf{Block boundaries.} A particle whose block is complete waits until every
particle has completed it; resampling is considered at these boundaries.}

\subsection{Sequential Monte Carlo decoding}
\label{sec:tilt}

\paragraph{The sampler.}
The sampler uses the decoder as its proposal\fixcut{, the distribution from which new
particle states are drawn,} and maintains \fixcut{a population of }$N$
particles\fixcut{. Particle}\fixnew{; particle} $i$ is a partially decoded sequence
$x^{(i)}_t$ with \fixcut{a }weight $U^{(i)}_t$\fixcut{; the particle index is written as
a superscript,} and either index is omitted when \fixcut{it is }unambiguous\fixcut{ from
context}. After the last block\fixcut{ a single particle is returned; which one is a
choice of selection rule, varied in Section 5.3}\fixnew{, a selection rule returns an
answer from the final population; the rule is varied in Section~\ref{sec:selectors}}. We
write $\tau = (x_T, \dots, x_0)$ for a complete decoding path.

\paragraph{Kernel and potential.}
For a step from $x_t$ to $x_{t-1}$, the \emph{transition kernel} $K_t(x_t, x_{t-1})$ is the
probability that one decoding step moves $x_t$ to $x_{t-1}$, including the token draws,
the commit selection, and the copying of unmasked positions. The step also carries a \emph{potential} $G_{t-1}(x_t, x_{t-1}) > 0$, a score of the tokens committed at that step. SR-SMC \citep{luo2026srsmc} takes the potential to be the product of the
confidences of the committed tokens,
\begin{equation}
  G_{t-1}(x_t, x_{t-1}) = \prod_{j \in \mathcal{S}_t} c_t(j) ,
  \label{eq:G}
\end{equation}
so the decoder supplies its own incremental potential. \fixcut{Accumulated over steps,
this is the cumulative confidence score of the trajectory, not of any single step,
which is what separates the weight of SR-SMC from the per-step confidence that chooses the
commit set.}\fixnew{Accumulated over steps this is the trajectory's cumulative confidence,
distinct from the per-step confidence that chooses the commit set.} The potential need not
be the decoder's confidence: Section~\ref{sec:experiments} also uses an external process
reward model, and nothing below depends on where $G$ comes from.

% Cut 2026-09-19: the outcome-vs-process history (cobbe2021gsm8k, uesato2022process,
% lightman2024verify) goes to Related work; the model name and the exact potential [C]
% go to Section 5.1. Citations inside \fixcut are plain text (\sout cannot hold \cite).

%% \paragraph{The sampler.}
%% The sampler uses the decoder as its proposal, the distribution from which new particle states are drawn, and maintains a population of $N$ particles. Particle $i$ is a partially decoded sequence $x^{(i)}_t$ with a weight $U^{(i)}_t$; the particle index is written as a superscript, and either index is dropped when it is clear from context. Each particle is advanced with
%% $K_t$ by running one decoding step, its weight is updated using $G_{t-1}$, and the
%% population is resampled when the effective sample size of the weights falls below $N/2$.
%% After the last block a single particle is returned; which one is a choice of
%% selector, varied in Section~\ref{sec:selectors}.
%% We write $\tau = (x_T, \dots, x_0)$ for a
%% complete decoding path.
%% 
%% 
% ^^ MOVED/REMOVED annotation #26: moved to the start of 2.2, 2026-09-24
% Moved 2026-09-19 to the end of 2.2, extended by one sentence.

\paragraph{Weights.}
Between resampling events, the weight of a particle is built from the incremental
weights of its steps:
\begin{equation}
  U_{t-1} = U_t\, G_{t-1}(x_t, x_{t-1}), \qquad U_T = 1 .
  \label{eq:sr-weight}
\end{equation}
This is the standard incremental weight of sequential Monte Carlo for the proposal
$q_{t-1} = K_t$ \citep{doucet2001sequential}, and it is the weight of SR-SMC.
% The next sentence moved here from the former 2.2 (unchanged).
Weight degeneracy is monitored with the normalized weights
$w_i = U^{(i)} / \sum_{k} U^{(k)}$ and the effective sample size
$\mathrm{ESS}(w) = 1 / \sum_i w_i^{2} \in [1, N]$\fixcut{, equivalently
$(\sum_i U^{(i)})^{2} / \sum_i (U^{(i)})^{2}$ on the unnormalized weights}, evaluated at
block boundaries; the sampler resamples when $\mathrm{ESS} < N/2$\fixcut{; we call this
test the ESS gate; when it is met, resampling is triggered. At a resampling event the
weights are replaced by the output of the resampler (Section 2.3): reset to $1/N$ under
multinomial resampling, or replaced by Chopthin's unequal output weights}\fixnew{, which
we call the ESS gate}.

\paragraph{What the weighted particles represent.}
\fixcut{Let $Q(\tau)$ denote the decoder path distribution and $C(\tau)$ the accumulated
potential, the product of the $G_{t-1}$ along the path. Without resampling, particles are
drawn from $Q$ and weighted by $C$, yielding a self-normalized importance-sampling
approximation to a target with unnormalized density proportional to $Q(\tau)\,C(\tau)$. A
resampler that is unbiased in the sense of Section 2.3 preserves this weighted empirical
measure in conditional expectation (Douc et al., 2005; Ahmadi et al., 2026). Thus changing
the resampler changes the finite-particle representation of the target, not the target
itself. This concerns the resampling operator alone: the weight floor of Section 4.2
explicitly modifies the weighted measure, and Section 4.2 bounds the change it makes.}
\fixnew{Let $Q(\tau)$ be the decoder path distribution and $C(\tau)$ the accumulated
potential, the product of the $G_{t-1}$ along the path. Without resampling, particles drawn
from $Q$ and weighted by $C$ form a self-normalized importance-sampling approximation to
the target with unnormalized density $Q(\tau)\,C(\tau)$; a resampler that is unbiased in
the sense of Section~\ref{sec:resamplers} preserves this weighted empirical measure in
conditional expectation \citep{douc2005comparison, ahmadi2026ccps}, so changing the
resampler changes the finite-particle representation of the target, not the target. The weight floor of Section~\ref{sec:cap}, by contrast, explicitly modifies the measure, and
Section~\ref{sec:cap} bounds the change.} 

%% \paragraph{Three objects.}
%% We distinguish the decoding \emph{path} $\tau$, also called a trajectory, the final \emph{output} $x_0$,
%% and the \emph{answer} the grader extracts from it; several paths may produce the same
%% output, and several outputs may carry the same answer. Counts of distinct trajectories reported below are counts of distinct outputs $x_0$.
% ^^ MOVED/REMOVED annotation #28: shortened and moved to Section 5, 2026-09-24

% The selector sentence moved to Section 5 on 2026-09-20.

\subsection{Resampling operators}
\label{sec:resamplers}
\label{sec:interface}

\fixnew{Every resampler in this paper takes the same input and returns the same kind of
output. It receives the normalized weights $w_{1:N}$ and returns $N$ ancestors $A_{1:N}$
and $N$ new weights $w^{+}_{1:N}$: output particle $k$ continues the state of input
particle $A_k$ and starts with weight $w^{+}_k$. The number of offspring of input particle
$i$ is $O_i = |\{k : A_k = i\}|$; $O_i = 0$ means it has no descendants.}
We
consider two existing resampling operators.

\paragraph{Multinomial.}
The ancestors are drawn independently, $A_k \sim \mathrm{Categorical}(w_{1:N})$ for
$k = 1, \dots, N$, and every returned weight is $w^{+}_k = 1/N$,
so particle $i$ receives $\mathrm{Binomial}(N, w_i)$ offspring\fixnew{; this is the resampler of SR-SMC}. \fixnew{Stratified, residual and systematic schemes reduce the variance of the
offspring counts but also return equal weights \citep{douc2005comparison};
Section~\ref{sec:ablation} tests systematic resampling.}

\paragraph{Chopthin.}
% The offspring function and its label eq:h moved here from Appendix A.1 on 2026-09-20;
% the equation is verbatim. The appendix keeps the integer counts, the systematic draws
% and the weight correction.
Chopthin \citep{gandy2016chopthin} takes a ratio bound $\eta \ge 4$ and computes a
threshold $a > 0$ from the weights. The expected number of offspring of particle $i$ is
\begin{equation}
  h^{\eta}_a(w_i) =
  \begin{cases}
    w_i / a, & w_i < a \quad \text{(thin: kept with probability } w_i/a\text{, weight set to } a\text{)},\\
    1, & a \le w_i < \eta a / 2 \quad \text{(keep: unchanged)},\\
    2 w_i / (\eta a), & w_i \ge \eta a / 2 \quad \text{(chop: split into copies of equal weight)},
  \end{cases}
  \label{eq:h}
\end{equation}
 where the threshold $a$ is chosen so that the expected
offspring counts sum to $N$; the coupled procedure returns exactly $N$ particles.
\fixcut{Light particles are thinned, particles of intermediate weight pass through with
their own weight, and heavy particles are chopped into copies of equal weight.}
Every returned weight lies in $[a, \eta a]$, so the ratio of the largest to the smallest
returned weight is at most $\eta$, and the weights need not be equalized. Details of the integer allocation and the weight correction
are given in Appendix~\ref{app:essblind}.

\fixnew{Multinomial resampling and Chopthin are both unbiased} in the sense that the
expected total offspring weight attributable to particle $i$
equals $w_i$, that is $\mathbb{E}\big[\sum_{k : A_k = i} w^{+}_k \,\big|\, w\big] = w_i$ \citep{douc2005comparison, gandy2016chopthin}, and therefore both preserve the target of Section~\ref{sec:tilt}. \fixnew{The rest of the paper varies two quantities: the
resampling operator, and the weights it is given.}

% =====================================================================
% Section 3 -- Motivation. Rebuilt 2026-09-17 on the three-move structure of the CCPS
% paper (ahmadi2026ccps, Sec. 3), which a reviewer singled out: (1) the two axes are
% different objectives, (2) two named costs with one underlying cause, (3) a design
% objective that derives the method as the weakest condition that suffices. The sequel
% here is that one weakest-sufficient-condition is not enough: Chopthin bounds the
% resampler's OUTPUT, and a particle can arrive so far behind that a gentle resampler
% still thins it away. That second gap is what the floor closes.
% =====================================================================
\section{Motivation: what a resampling event discards}
\label{sec:motivation}

\fixnew{Under SR-SMC only $5.2$ of $16$ lineages survive on GSM8K with LLaDA-1.5 and
$1.1$ with Dream-7B, and between $9.1$ and $11.7$ of $16$ particles lose all offspring
per resampling event (Table~\ref{tab:ablation2}, Section~\ref{sec:ablation}).}

% 2026-09-25: both subsection headings converted to run-in headings (reviewer); labels kept.
\paragraph{ESS and lineage diversity.}
\label{sec:ess-blind}

\fixnew{To count lineages, each particle carries a \emph{root} $r(i)$: the index of the initial
particle it descends from (two initial particles get different roots even if their
sequences coincide). The number of surviving lineages $R$ is the number of distinct roots
in the population.} \fixcut{A function of $w$ alone, such as the ESS, does not depend on
$r$. For every $w$ and every $R \in \{1, \dots, N\}$ there is a root map with exactly $R$
distinct roots. So a trigger that looks only at the current weights cannot tell how many
lineages survive. The following proposition quantifies what such a trigger misses, for the
two resamplers of Section 2.3, when they are invoked.}\fixnew{The ESS is a function of
$w$ alone, and for every $w$ and every $R \in \{1, \dots, N\}$ there is a root map with
exactly $R$ distinct roots, so a trigger that depends only on the weights cannot tell how
many lineages survive. Proposition~\ref{prop:essblind} quantifies what it misses for the
two resamplers of Section~\ref{sec:resamplers}.}

\begin{proposition}[ESS-blindness]
\label{prop:essblind}
Let $\mathcal{P} = (w, r)$ have positive weights, let $\mathcal{P}^{+} = (w^{+}, r^{+})$
be the population after one resampling event applied to it, and let $R^{+}$ be its number
of lineages. For a root $\rho$ present in $\mathcal{P}$, let $w_\rho = \sum_{i : r(i) = \rho} w_i$
be the total weight of its lineage.
\begin{enumerate}
\item[(i)] After multinomial resampling, $\mathrm{ESS}(w^{+}) = N$ always, while
\[
  \mathbb{E}[R^{+} \mid \mathcal{P}] = \sum_{\rho} \big(1 - (1 - w_\rho)^{N}\big),
\]
the sum running over the $R$ roots present in $\mathcal{P}$. For uniform weights and $R = N$
this is $N\big(1 - (1 - 1/N)^{N}\big)$: the expected surviving fraction
$1 - (1 - 1/N)^{N}$ decreases with $N$ to $1 - e^{-1} \approx 0.632$, and equals $0.644$
at $N = 16$, that is, $10.3$ of $16$ lineages.
\item[(ii)] If $\max_i w_i / \min_i w_i \le \eta / 2$, then Chopthin returns
$\mathcal{P}$ unchanged up to the order of the particles: every particle survives with
its own weight, so $w^{+} = w$ and $R^{+} = R$. Uniform weights always satisfy this
condition.
\end{enumerate}
\end{proposition}

The proof is in Appendix~\ref{app:essblind}.

\begin{remark}
Parts (i) and (ii) describe two populations with the same weight statistics before and
after the event (both start and end at $\mathrm{ESS} = N$) whose genealogies differ by
about a third of the lineages for any $N \ge 4$ ($36\%$ at $N = 16$). The ESS does not
distinguish them; a root count can. The loss in (i) is specific to multinomial
resampling\fixcut{, which is the resampler of SR-SMC; systematic or stratified resampling
on exactly uniform weights gives every particle one offspring and keeps all lineages (Douc
et al., 2005)}\fixnew{; on exactly uniform weights systematic or stratified resampling
keeps every lineage \citep{douc2005comparison}, and Section~\ref{sec:ablation} tests
systematic resampling}.

%% The $36\%$ figure assumes uniform weights, and is therefore a floor rather than an
%% estimate. Measured at the resampling events of the runs in Section~\ref{sec:ablation},
%% multinomial resampling leaves $9.1$ to $11.7$ of $16$ particles without offspring per
%% event, that is $57\%$ to $73\%$, between one and a half and two times the uniform-weight
%% expectation. The gap is the weights: the more concentrated they are when the gate fires,
%% the more the event discards.
% ^^ REMOVED annotation #39: deleted; its number moved to the Section 3 opening, 2026-09-24

\end{remark}

\paragraph{Requirements on the resampling operator.}
\label{sec:twocosts}
\label{sec:objective}
\fixnew{Flattening the weights removes degeneracy at a cost at both ends of the weight
distribution: a particle holding less than a $1/N$ share is likely to receive no
offspring, deleted on the basis of a weight that scores an unfinished prefix, while a
heavy particle is copied into descendants that share a prefix, so an event never increases
the number of distinct prefixes and usually decreases it \citep{doucet2001sequential};
Figure~\ref{fig:overview} shows both at one boundary. Neither cost is necessary: a resampling step need only leave
the target unchanged, return $N$ particles and keep the weights from concentrating.
Equalizing them is stronger than necessary: by Lemma~\ref{lem:essfloor} a bound of $\eta$
on the ratio of the largest to the smallest weight already implies
$\mathrm{ESS} \ge N/2$, and Chopthin enforces exactly this bound on the weights it returns
(Section~\ref{sec:chopthin}). That bound constrains only the output of the step; a particle whose weight is far below the maximum when the resampler is called is removed with probability approaching one under any such
rule, and the weight floor bounds this input-side gap (Section~\ref{sec:cap}). Neither
bound depends on the origin of the weights, so both apply when the potential is a reward
model.}

\section{LP-SMC: Chopthin resampling with a weight floor}
\label{sec:method}

LP-SMC denotes the population-management step only: Chopthin
resampling with the weight floor of Section~\ref{sec:cap}.
The guidance signal and the selection rule are varied separately in
Section~\ref{sec:experiments}. %

\paragraph{At a block boundary.} (i)~the running weights $U$ have been updated by
\eqref{eq:sr-weight} through the block; (ii)~the floor of
Section~\ref{sec:cap} is applied, giving $\tilde U$ and $\tilde w$; (iii)~if
$\mathrm{ESS}(\tilde w) < N/2$, Chopthin (Section~\ref{sec:chopthin}) is applied to
$\tilde w$ and each particle continues from its ancestor with weight $w^{+}$;
(iv)~otherwise every particle continues with weight $\tilde U$. Decoding then resumes.

\subsection{Chopthin resampling in diffusion SMC}
\label{sec:chopthin}
We modify diffusion SMC only at the
population-management stage. The decoder transition $K_t$, the potential $G_{t-1}$, the
block schedule and the $\mathrm{ESS} < N/2$ trigger are unchanged and no model
evaluations are added. The population-management step therefore applies whether the boundary weights come
from accumulated decoder confidence or an external reward model.
We replace the resampling step by Chopthin
\citep{gandy2016chopthin} with carried weights \citep{ahmadi2026ccps}:
\begin{equation}
  (A_{1:N}, w^{+}_{1:N}) = \mathrm{Chopthin}(w_{1:N}, \eta), \qquad
  x^{(k)} \leftarrow x^{(A_k)}, \qquad U^{(k)} \leftarrow w^{+}_k .
  \label{eq:chopthin}
\end{equation}
The assignments are
simultaneous: output particle $k$ continues the full state of input particle $A_k$,
including its root label and cached decoder state. The
returned weights are defined up to a common factor; under internal guidance they are
carried into the next block instead of being reset; under external guidance the weight is
rebuilt from the reward model at each boundary (Section~\ref{sec:experiments}, Setup).
Since
every returned weight lies in $[a, \eta a]$,
Lemma~\ref{lem:essfloor} gives $\mathrm{ESS}(w^{+}) \ge 4 \eta N / (\eta + 1)^{2}$,
which tightens the bound of \citet{gandy2016chopthin} by a
constant. We use $\eta = 3 + \sqrt{8}$ throughout, for which the
bound is exactly $N/2$: an event cannot leave the population below the
threshold that triggered it. This is a guarantee on the weights, not on the number of
lineages. The quantities have separate roles: the potential sets how far the weights
spread, $\eta$ sets the spread of the resampler's output, the floor of
Section~\ref{sec:cap} sets how much it tolerates on its input, and the ESS gate sets when
the resampler is called at all.

\subsection{Flooring the input weights}
\label{sec:cap}
Chopthin bounds the ratio of the weights it \emph{returns}, not of the weights it is \emph{given}, and a particle whose weight is far below the maximum
is removed with high probability regardless of the output bound. We therefore bound the
input ratio as well. At a block boundary let $U_{\max} = \max_k U^{(k)}$;
the floored weights are
\begin{equation}
  \tilde U^{(i)} = \max\!\big(U^{(i)},\, e^{-\kappa} U_{\max}\big), \quad
  \tilde w_i = \tilde U^{(i)} \Big/ \sum_j \tilde U^{(j)}, \quad \kappa > 0 .
  \label{eq:cap}
\end{equation}
The gate tests $\mathrm{ESS}(\tilde w) < N/2$ and the resampler receives
$\tilde w$. The
floor is a deterministic map that needs no model evaluation and applies to any resampler;
it leaves the maximum unchanged and raises the lowest weights, the opposite end of the
distribution from Chopthin's chop step.

Under internal guidance the floored or post-resampling weights are carried forward:
between boundaries $U_{t-1} = \tilde U_t\, G_{t-1}$, with the floor re-applied at every boundary.
Under external guidance weights are rebuilt at every boundary, so neither the floor-induced increase
nor the returned Chopthin weights persist; this is a boundary-wise reweighting, not the accumulated
potential of \eqref{eq:sr-weight}, so we treat the externally guided setting as a robustness check
and not as the same Feynman--Kac target \citep{delmoral2004feynman}.

After~\eqref{eq:cap} the ratio of $\tilde U$ is at most $e^{\kappa}$, so
Lemma~\ref{lem:essfloor} applies to the weights that the gate and the resampler receive:
$\mathrm{ESS} \ge 4 e^{\kappa} N / (e^{\kappa} + 1)^{2}$. The ESS gate imposes a lower bound on $\kappa$; we
then choose $\kappa = \ln N$ to keep the floor non-vacuous.

\emph{Gate compatibility.} Once this lower
bound exceeds the gate threshold, resampling is never triggered; the sampler resamples only
when
\begin{equation}
  \kappa \;>\; \kappa^{\star} := \ln\big(3 + \sqrt{8}\big) \approx 1.763 .
  \label{eq:kappastar}
\end{equation}
The corresponding ratio $e^{\kappa^{\star}} = 3+\sqrt{8}$ equals the Chopthin ratio
bound $\eta$ of Section~\ref{sec:chopthin}; both are chosen so that the worst admissible weight
profile has $\mathrm{ESS} = N/2$. %

\emph{Choice of $\kappa$.} Conversely, as $\kappa$ grows
the bound falls back to zero and the floor has no effect. We use
$\kappa = \ln N$, a floor at $1/N$ of the maximum
weight, which satisfies the gate condition for $N \ge 6$ (at $N = 16$,
$\kappa = 2.77$ against $\kappa^{\star} = 1.76$). Under this floor a particle's normalized
weight is at least $1/(N(N-1)+1)$, so a multinomial ancestor draw keeps it with
probability at least $1 - (1 - 1/(N(N-1)+1))^{N}$, $0.064$ at $N = 16$
(Appendix~\ref{app:floor}); without the floor no positive lower bound exists. Empirically, in the GSM8K lineage-risk analysis, no correctly answered problem under either floored LP-SMC configuration reaches
the corresponding deletion probability of $0.936$ (Table~\ref{tab:lineagerisk},
Appendix~\ref{app:lineagerisk}).

Raising low weights changes what the resampler is unbiased for. Resampling remains
unbiased, but for the floored weights $\tilde w$ rather than the original $w$. The change is
confined to the particles the floor actually touches: a particle already within $\kappa$ nats
of the maximum keeps its weight exactly, and only particles below that threshold are raised,
to $e^{-\kappa}$ of the maximum. Under internal guidance the raised value persists until the
next resampling event.

The size of the change is bounded. At a single boundary the floor can shift at most
$(N-1)/(2N-1)$ of the normalized mass onto floored particles, which is $0.48$ at $N = 16$,
and that worst case arises only when $N-1$ of the $N$ particles sit on the floor at once.
In the measured runs the mass on floored particles is about an order of
magnitude smaller (Appendix~\ref{app:floor}). The floor is therefore an explicit and bounded
change to the weights, in the same family as the weight flattening used by particle methods
for reasoning \citep{giannone2025epf}.

\section{Experiments}
\label{sec:experiments}

\paragraph{Setup.}
The decoders are
LLaDA-1.5 \citep{llada2025} and Dream-7B \citep{dream2025}. Both run with generation length
$L = 256$, block size $32$, eight decoding steps per block, confidence-threshold commitment
at $0.9$, temperature $1.0$ and $N = 16$ particles. Resampling is gated at
$\mathrm{ESS} < N/2$; Chopthin uses $\eta = 3 + \sqrt{8}$ and the floor $\kappa = \ln N$.
Under internal guidance the weight accumulates by \eqref{eq:sr-weight}; under external guidance it is rebuilt at every boundary from the
log-odds of the mean step score of Qwen2.5-Math-PRM-7B \citep{yang2024qwen25math} on the
current prefix, as in \citet{puri2025reasoningsmc} (Appendix~\ref{app:prmscore} gives the exact definition). Benchmarks are MATH-500 \citep{hendrycks2021math,lightman2024verify} (4-shot),
GSM8K \citep{cobbe2021gsm8k} (5-shot), HumanEval \citep{chen2021humaneval} (0-shot) and
MBPP \citep{austin2021mbpp} (3-shot), on the full test set of each; grading and shot
conventions are in Appendix~\ref{app:protocol}. The population-management operations
introduce no direct decoder or reward-model evaluations; Appendix~\ref{app:protocol}
reports measured end-to-end NFE and wall time.

\subsection{Main comparison}
\label{sec:guidance}

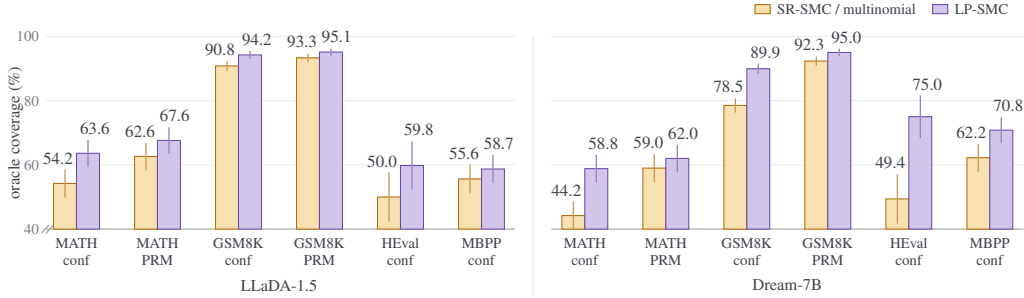
\begin{figure}[t]
\centering
\definecolor{ocB}{RGB}{176,110,10}   \definecolor{ocBF}{RGB}{248,224,176}  % baseline: gold/sand (was drmM/drmMF)
\definecolor{ocL}{RGB}{109,76,166}   \definecolor{ocLF}{RGB}{209,197,234}  % LP-SMC: violet (was drmC/drmCF)
\definecolor{ocSep}{RGB}{226,226,226}
\definecolor{ocInk}{RGB}{62,66,74}
\definecolor{ocMut}{RGB}{120,124,132}
\begin{tikzpicture}[
  font=\footnotesize,
  val/.style={font=\fontsize{6.4}{7.4}\selectfont, text=ocInk, inner sep=0.5pt, anchor=south},
  ytk/.style={font=\fontsize{5.6}{6.6}\selectfont, text=ocMut, anchor=east, inner sep=1.3pt},
  xtk/.style={font=\fontsize{5.6}{6.6}\selectfont, text=ocInk, align=center, inner sep=1pt},
  mdl/.style={font=\fontsize{6.2}{7}\selectfont, text=ocInk, inner sep=1pt},
  ylb/.style={font=\fontsize{6.2}{7}\selectfont, text=ocInk, rotate=90},
  lgd/.style={font=\fontsize{5.6}{6.6}\selectfont, text=ocInk, anchor=west, inner sep=1pt},
  grid/.style={draw=black!8, line width=0.3pt},
  axis/.style={draw=black!35, line width=0.4pt},
]
%% hairline splitting the two panels
  \draw[draw=ocSep, line width=0.4pt] (6.965,-0.880) -- (6.965,2.550);
%% gridlines first, so the bars sit on top of them
  \draw[grid] (0.520,0.000) -- (6.725,0.000);
  \draw[grid] (7.205,0.000) -- (13.450,0.000);
  \node[ytk] at (0.480,0.000) {40};
  \draw[grid] (0.520,0.850) -- (6.725,0.850);
  \draw[grid] (7.205,0.850) -- (13.450,0.850);
  \node[ytk] at (0.480,0.850) {60};
  \draw[grid] (0.520,1.700) -- (6.725,1.700);
  \draw[grid] (7.205,1.700) -- (13.450,1.700);
  \node[ytk] at (0.480,1.700) {80};
  \draw[grid] (0.520,2.550) -- (6.725,2.550);
  \draw[grid] (7.205,2.550) -- (13.450,2.550);
  \node[ytk] at (0.480,2.550) {100};
  \draw[axis] (0.520,0) -- (6.725,0);
  \draw[axis] (7.205,0) -- (13.450,0);
  \node[ylb] at (0.120,1.275) {oracle coverage (\%)};
%% axis break mark at the clipped baseline
  \draw[axis, line width=0.5pt] (0.460,-0.055) -- (0.560,0.055);
  \draw[axis, line width=0.5pt] (0.520,-0.055) -- (0.620,0.055);
  \fill[ocBF] (0.620,0) rectangle (0.924,0.604);
  \draw[ocB, line width=0.4pt] (0.620,0) rectangle (0.924,0.604);
  \draw[draw=ocB!70, line width=0.5pt] (0.772,0.418) -- (0.772,0.789);
  \fill[ocLF] (0.924,0) rectangle (1.228,1.003);
  \draw[ocL, line width=0.4pt] (0.924,0) rectangle (1.228,1.003);
  \draw[draw=ocL!70, line width=0.5pt] (1.076,0.824) -- (1.076,1.182);
  \node[val] at (0.679,0.839) {54.2};
  \node[val] at (1.168,1.232) {63.6};
  \node[xtk, anchor=north] at (0.924,-0.07) {MATH\\conf};
  \fill[ocBF] (1.692,0) rectangle (1.995,0.961);
  \draw[ocB, line width=0.4pt] (1.692,0) rectangle (1.995,0.961);
  \draw[draw=ocB!70, line width=0.5pt] (1.843,0.780) -- (1.843,1.141);
  \fill[ocLF] (1.995,0) rectangle (2.299,1.173);
  \draw[ocL, line width=0.4pt] (1.995,0) rectangle (2.299,1.173);
  \draw[draw=ocL!70, line width=0.5pt] (2.147,0.999) -- (2.147,1.347);
  \node[val] at (1.766,1.191) {62.6};
  \node[val] at (2.225,1.397) {67.6};
  \node[xtk, anchor=north] at (1.995,-0.07) {MATH\\PRM};
  \fill[ocBF] (2.763,0) rectangle (3.067,2.159);
  \draw[ocB, line width=0.4pt] (2.763,0) rectangle (3.067,2.159);
  \draw[draw=ocB!70, line width=0.5pt] (2.915,2.093) -- (2.915,2.225);
  \fill[ocLF] (3.067,0) rectangle (3.371,2.304);
  \draw[ocL, line width=0.4pt] (3.067,0) rectangle (3.371,2.304);
  \draw[draw=ocL!70, line width=0.5pt] (3.219,2.250) -- (3.219,2.357);
  \node[val] at (2.822,2.275) {90.8};
  \node[val] at (3.311,2.407) {94.2};
  \node[xtk, anchor=north] at (3.067,-0.07) {GSM8K\\conf};
  \fill[ocBF] (3.835,0) rectangle (4.138,2.265);
  \draw[ocB, line width=0.4pt] (3.835,0) rectangle (4.138,2.265);
  \draw[draw=ocB!70, line width=0.5pt] (3.987,2.208) -- (3.987,2.323);
  \fill[ocLF] (4.138,0) rectangle (4.442,2.342);
  \draw[ocL, line width=0.4pt] (4.138,0) rectangle (4.442,2.342);
  \draw[draw=ocL!70, line width=0.5pt] (4.290,2.292) -- (4.290,2.391);
  \node[val] at (3.909,2.373) {93.3};
  \node[val] at (4.368,2.441) {95.1};
  \node[xtk, anchor=north] at (4.138,-0.07) {GSM8K\\PRM};
  \fill[ocBF] (4.906,0) rectangle (5.210,0.425);
  \draw[ocB, line width=0.4pt] (4.906,0) rectangle (5.210,0.425);
  \draw[draw=ocB!70, line width=0.5pt] (5.058,0.100) -- (5.058,0.750);
  \fill[ocLF] (5.210,0) rectangle (5.514,0.841);
  \draw[ocL, line width=0.4pt] (5.210,0) rectangle (5.514,0.841);
  \draw[draw=ocL!70, line width=0.5pt] (5.362,0.523) -- (5.362,1.160);
  \node[val] at (4.966,0.800) {50.0};
  \node[val] at (5.454,1.210) {59.8};
  \node[xtk, anchor=north] at (5.210,-0.07) {HEval\\conf};
  \fill[ocBF] (5.978,0) rectangle (6.282,0.663);
  \draw[ocB, line width=0.4pt] (5.978,0) rectangle (6.282,0.663);
  \draw[draw=ocB!70, line width=0.5pt] (6.130,0.474) -- (6.130,0.852);
  \fill[ocLF] (6.282,0) rectangle (6.585,0.795);
  \draw[ocL, line width=0.4pt] (6.282,0) rectangle (6.585,0.795);
  \draw[draw=ocL!70, line width=0.5pt] (6.433,0.608) -- (6.433,0.982);
  \node[val] at (6.052,0.902) {55.6};
  \node[val] at (6.511,1.032) {58.7};
  \node[xtk, anchor=north] at (6.282,-0.07) {MBPP\\conf};
  \fill[ocBF] (7.345,0) rectangle (7.648,0.179);
  \draw[ocB, line width=0.4pt] (7.345,0) rectangle (7.648,0.179);
  \draw[draw=ocB!70, line width=0.5pt] (7.497,0.000) -- (7.497,0.364);
  \fill[ocLF] (7.648,0) rectangle (7.952,0.799);
  \draw[ocL, line width=0.4pt] (7.648,0) rectangle (7.952,0.799);
  \draw[draw=ocL!70, line width=0.5pt] (7.800,0.616) -- (7.800,0.982);
  \node[val] at (7.404,0.414) {44.2};
  \node[val] at (7.893,1.032) {58.8};
  \node[xtk, anchor=north] at (7.648,-0.07) {MATH\\conf};
  \fill[ocBF] (8.416,0) rectangle (8.720,0.807);
  \draw[ocB, line width=0.4pt] (8.416,0) rectangle (8.720,0.807);
  \draw[draw=ocB!70, line width=0.5pt] (8.568,0.624) -- (8.568,0.991);
  \fill[ocLF] (8.720,0) rectangle (9.024,0.935);
  \draw[ocL, line width=0.4pt] (8.720,0) rectangle (9.024,0.935);
  \draw[draw=ocL!70, line width=0.5pt] (8.872,0.754) -- (8.872,1.116);
  \node[val] at (8.491,1.041) {59.0};
  \node[val] at (8.949,1.166) {62.0};
  \node[xtk, anchor=north] at (8.720,-0.07) {MATH\\PRM};
  \fill[ocBF] (9.488,0) rectangle (9.792,1.636);
  \draw[ocB, line width=0.4pt] (9.488,0) rectangle (9.792,1.636);
  \draw[draw=ocB!70, line width=0.5pt] (9.640,1.542) -- (9.640,1.730);
  \fill[ocLF] (9.792,0) rectangle (10.095,2.121);
  \draw[ocL, line width=0.4pt] (9.792,0) rectangle (10.095,2.121);
  \draw[draw=ocL!70, line width=0.5pt] (9.943,2.052) -- (9.943,2.190);
  \node[val] at (9.547,1.780) {78.5};
  \node[val] at (10.036,2.240) {89.9};
  \node[xtk, anchor=north] at (9.792,-0.07) {GSM8K\\conf};
  \fill[ocBF] (10.559,0) rectangle (10.863,2.223);
  \draw[ocB, line width=0.4pt] (10.559,0) rectangle (10.863,2.223);
  \draw[draw=ocB!70, line width=0.5pt] (10.711,2.162) -- (10.711,2.284);
  \fill[ocLF] (10.863,0) rectangle (11.167,2.337);
  \draw[ocL, line width=0.4pt] (10.863,0) rectangle (11.167,2.337);
  \draw[draw=ocL!70, line width=0.5pt] (11.015,2.288) -- (11.015,2.387);
  \node[val] at (10.619,2.334) {92.3};
  \node[val] at (11.108,2.437) {95.0};
  \node[xtk, anchor=north] at (10.863,-0.07) {GSM8K\\PRM};
  \fill[ocBF] (11.631,0) rectangle (11.935,0.399);
  \draw[ocB, line width=0.4pt] (11.631,0) rectangle (11.935,0.399);
  \draw[draw=ocB!70, line width=0.5pt] (11.783,0.074) -- (11.783,0.725);
  \fill[ocLF] (11.935,0) rectangle (12.238,1.487);
  \draw[ocL, line width=0.4pt] (11.935,0) rectangle (12.238,1.487);
  \draw[draw=ocL!70, line width=0.5pt] (12.087,1.206) -- (12.087,1.769);
  \node[val] at (11.690,0.775) {49.4};
  \node[val] at (12.179,1.819) {75.0};
  \node[xtk, anchor=north] at (11.935,-0.07) {HEval\\conf};
  \fill[ocBF] (12.703,0) rectangle (13.006,0.944);
  \draw[ocB, line width=0.4pt] (12.703,0) rectangle (13.006,0.944);
  \draw[draw=ocB!70, line width=0.5pt] (12.854,0.760) -- (12.854,1.127);
  \fill[ocLF] (13.006,0) rectangle (13.310,1.309);
  \draw[ocL, line width=0.4pt] (13.006,0) rectangle (13.310,1.309);
  \draw[draw=ocL!70, line width=0.5pt] (13.158,1.136) -- (13.158,1.482);
  \node[val] at (12.762,1.177) {62.2};
  \node[val] at (13.251,1.532) {70.8};
  \node[xtk, anchor=north] at (13.006,-0.07) {MBPP\\conf};
  \node[mdl] at (3.603,-0.74) {LLaDA-1.5};
  \node[mdl] at (10.327,-0.74) {Dream-7B};
  \fill[ocBF] (9.900,2.850) rectangle (10.100,2.970);
  \draw[ocB, line width=0.4pt] (9.900,2.850) rectangle (10.100,2.970);
  \node[lgd] at (10.150,2.910) {SR-SMC / multinomial};
  \fill[ocLF] (12.250,2.850) rectangle (12.450,2.970);
  \draw[ocL, line width=0.4pt] (12.250,2.850) rectangle (12.450,2.970);
  \node[lgd] at (12.500,2.910) {LP-SMC};
\end{tikzpicture}
\caption[Oracle coverage, the fraction of problems with at least one correct particle.]{%
Oracle coverage, the fraction of problems with at least one correct particle
($N = 16$, full test sets), under the decoder's confidence (conf) or the external PRM
in the weight. LP-SMC is higher in all twelve settings. Bars: unpaired $95\%$ binomial
intervals; paired \fixcut{discordant counts and bootstrap CIs on the differences are in}\fixnew{counts and
bootstrap intervals:} Appendix~\ref{app:paired}. The axis is clipped at $40$\fixcut{ (break mark at the
origin), so bar length is not proportional to coverage}.}
\label{fig:oracle}
\end{figure}

Table~\ref{tab:main} is the primary controlled comparison. Within each pair the
model, the guidance signal, the selection rule and the problems are held fixed
and only the population step changes, multinomial resampling against LP-SMC, so a
difference between the two rows cannot arise from the selection rule.
LP-SMC has the higher accuracy in eleven of the twelve accuracy cells, under the decoder's confidence on
all four benchmarks for both models and under the external reward model on the
mathematics benchmarks for LLaDA-1.5; the exception is Dream-7B on MATH-500 under
external guidance, $49.2$ against $48.2$. Coverage is higher under LP-SMC in all twelve
settings (Figure~\ref{fig:oracle}), and the coverage gains are also the better supported: of the $95\%$ bootstrap intervals on the paired
differences (Appendix~\ref{app:paired}), six of the twelve accuracy intervals include zero and eleven of the
twelve coverage intervals exclude it, and none favours the baseline.

%

% Figure: oracle coverage (generated by make_fig_oracle.py; edit the script, not this file).
% Needs \usepackage{tikz} only -- same requirement as fig_overview.tex.

\begin{table}[t]
\caption[End-to-end accuracy under reward-model selection.]{End-to-end accuracy under
reward-model selection (Qwen2.5-Math-PRM-7B \citep{yang2024qwen25math} on mathematics,
AceCodeRM-7B \citep{zeng2025acecoder} on code), $N{=}16$, $L{=}256$, paired problems, full test sets. Internal guidance: the decoder's confidence in the weight; External
PRM: the reward model in the weight. n.a.: no step-level code reward model
meeting our criteria is available, and AceCodeRM-7B, an outcome model, can rank finished
programs but not weight a prefix (Appendix~\ref{app:coderm}). Paired tests:
Appendix~\ref{app:paired}.}
\label{tab:main}
\begin{center}
\small
\setlength{\tabcolsep}{4.5pt}
\begin{tabular}{lllcccc}
\toprule
Model & Guidance & Population management & MATH-500 & GSM8K & HumanEval & MBPP \\
\midrule
\multirow{4}{*}{LLaDA-1.5} & \multirow{2}{*}{Internal}     & SR-SMC          & 49.0 & 88.2 & 48.2 & 49.8 \\
                           &                               & LP-SMC          & \textbf{52.4} & \textbf{90.5} & \textbf{56.1} & \textbf{50.6} \\
\cmidrule(lr){2-7}
                           & \multirow{2}{*}{External PRM} & Multinomial SMC & 53.8 & 90.4 & n.a. & n.a. \\
                           &                               & LP-SMC          & \textbf{55.0} & \textbf{90.8} & n.a. & n.a. \\
\midrule
\multirow{4}{*}{Dream-7B}  & \multirow{2}{*}{Internal}     & SR-SMC          & 41.8 & 78.2 & 48.8 & 59.8 \\
                           &                               & LP-SMC          & \textbf{46.8} & \textbf{86.7} & \textbf{67.1} & \textbf{62.0} \\
\cmidrule(lr){2-7}
                           & \multirow{2}{*}{External PRM} & Multinomial SMC & \textbf{49.2} & 90.8 & n.a. & n.a. \\
                           &                               & LP-SMC          & 48.2 & \textbf{91.7} & n.a. & n.a. \\
\bottomrule
\end{tabular}
\end{center}
\end{table}

\subsection{Where the gain comes from}
\label{sec:ablation}

Table~\ref{tab:ablation2} reports surviving lineages beside distinct outputs and
coverage, which bounds what any selection rule can reach.

Lineages increase at every step in all eight rows and coverage in seven; the exception is Dream on
GSM8K, where Chopthin alone costs a little coverage before the floor more than recovers it. Under multinomial resampling Dream is close to a
single surviving lineage on every benchmark, so the floor has more to recover there, and Chopthin
without the floor barely moves it. Distinct outputs rise with lineages in every row, so the extra lineages
carry different completions, not duplicates. Nor is the gain obtained by resampling less
often: LP-SMC triggers the gate as often or more often than multinomial resampling in all eight
settings while losing fewer than half as many particles per event, $4.01$ against $11.66$ on Dream
GSM8K, and its loss per event varies little across models and benchmarks where the multinomial loss
tracks how concentrated the weights happen to be.

\begin{table}[t]
\caption{Adding one component at a time under the decoder's confidence:
multinomial resampling (Mult.), Chopthin without the floor (Chop.), Chopthin
with the floor (Chop.+floor), which is LP-SMC. Lineages: mean number of founding particles still
represented at the end, out of 16; distinct outputs: mean number of distinct
$x_0$ per problem; coverage as in Figure~\ref{fig:oracle}.
Table~\ref{tab:capsep} crosses the floor with the resampler. Benchmark sizes are as in Table~\ref{tab:selectors}.}
\label{tab:ablation2}
\centering
\footnotesize
\setlength{\tabcolsep}{2.5pt}
\begin{tabular}{llccc@{\hskip 4pt}ccc@{\hskip 4pt}ccc}
\toprule
 & & \multicolumn{3}{c}{Surviving lineages} & \multicolumn{3}{c}{Distinct outputs} & \multicolumn{3}{c}{Coverage (\%)} \\
\cmidrule(lr){3-5} \cmidrule(lr){6-8} \cmidrule(lr){9-11}
Model & Benchmark & Mult. & Chop. & Chop.+floor & Mult. & Chop. & Chop.+floor & Mult. & Chop. & Chop.+floor \\
\midrule
\multirow{4}{*}{LLaDA-1.5}
 & MATH-500  & 2.71 & 3.93 & \textbf{6.98}  & 6.90 & 8.71 & \textbf{10.53} & 54.2 & 57.2 & \textbf{63.6} \\
 & GSM8K    &  5.21 &  7.52 & \textbf{10.04} & 7.80 & 9.65 & \textbf{11.28} &  90.8 &  92.9 & \textbf{94.2} \\
 & HumanEval &  2.96 &  4.58 & \textbf{7.79}  & 2.45 & 3.56 & \textbf{5.96}  &  50.0 &  54.9 & \textbf{59.8} \\
 & MBPP      & 10.36 & 11.76 & \textbf{13.45} & 5.88 & 6.89 & \textbf{8.37}  &  55.6 &  56.2 & \textbf{58.7} \\
\midrule
\multirow{4}{*}{Dream-7B}
 & MATH-500  & 1.55 & 2.18 & \textbf{5.37}  & 3.47 & 4.17 & \textbf{7.70}  & 44.2 & 47.4 & \textbf{58.8} \\
 & GSM8K    &  1.06 &  1.10 & \textbf{4.36}  & 1.23 & 1.33 & \textbf{5.43}  &  78.5 &  77.3 & \textbf{89.9} \\
 & HumanEval &  1.33 &  1.58 & \textbf{4.65}  & 3.11 & 3.83 & \textbf{8.11}  &  49.4 &  52.4 & \textbf{75.0} \\
 & MBPP      &  1.88 &  2.66 & \textbf{5.90}  & 4.96 & 5.92 & \textbf{9.62}  &  62.2 &  62.9 & \textbf{70.8} \\
\bottomrule
\end{tabular}
\end{table}

Table~\ref{tab:ablation2} adds the components in one order; crossing the ancestor draw with
the floor on GSM8K (Table~\ref{tab:capsep}, Appendix~\ref{app:floor}) tests whether either alone
would suffice. Systematic resampling raises surviving
lineages but leaves coverage at $90.8\%$, so drawing the ancestors more evenly is not a substitute
for bounding the weights that enter the draw, and Chopthin is ahead of it on every column of that
table, with and without the floor. The floor raises coverage on its own under
both draws, so it addresses a distinct failure mode; nevertheless no single-component
configuration reaches LP-SMC. Accuracy and
surviving lineages both rise monotonically with the depth of the floor on MATH-500
(Figure~\ref{fig:floorsweep}, Appendix~\ref{app:floor}); we keep $\kappa = \ln N$ throughout because
\eqref{eq:kappastar} admits it for every $N \ge 6$ with no tuning, although $1/8$ does better on
that benchmark. Among correctly answered LP-SMC runs, many correct lineages had a high hypothetical
deletion probability under multinomial resampling at their recorded weights
(Appendix~\ref{app:lineagerisk}).

\begin{table}[t]
\caption{Ancestor draw and floor separated: LLaDA-1.5 on GSM8K ($n{=}1319$),
decoder confidence, all configurations paired. Systematic resampling is the standard
variance-reduced draw. The six rows are a full factorial of ancestor draw
against weight floor. Lost per event: mean number of the 16 particles left
without offspring at a resampling event (lower is better).}
\label{tab:capsep}
\centering
\small
\begin{tabular}{lcccc}
\toprule
Configuration & Surviving lineages & Distinct outputs & Lost per event & Coverage (\%) \\
\midrule
Multinomial                                    & 5.21 & 7.80 & 9.14 & 90.8 \\
Multinomial + floor                            & 6.25 & 8.80 & 8.48 & 92.5 \\
\midrule
Systematic                                     & 6.21 & 8.43 & 7.98 & 90.8 \\
Systematic + floor                             & 7.62 & 9.48 & 6.97 & 93.6 \\
\midrule
Chopthin                                       & 7.52 & 9.65 & 5.41 & 92.9 \\
Chopthin + floor (LP-SMC)                          & \textbf{10.04} & \textbf{11.28} & \textbf{3.61} & \textbf{94.2} \\
\bottomrule
\end{tabular}
\end{table}

\subsection{Selection rules}
\label{sec:selectors}

Table~\ref{tab:selectors} (Appendix~\ref{app:selectors}) applies four selection rules ---
confidence, majority, distinct-output consensus, and the best-of-$N$ reward model of
Table~\ref{tab:main} \citep{brown2024monkeys,stroebl2024limits} --- to the same saved populations,
so selector quality is not confounded with population quality; each rule is defined in full there.

Coverage improves consistently; what varies is how much of it each rule converts into
accuracy. The one rule that favours the baseline is the decoder's own confidence, which by
construction undervalues the low-confidence lineages the floor exists to keep; the populations are
not weaker, and coverage rises in every setting regardless.
On Dream-7B the ordering is less uniform and the consensus rules split
(Table~\ref{tab:dreamselfull}, Appendix~\ref{app:dreamsel}); the two models are compared within a
table, not across tables, since Dream is a base checkpoint and LLaDA-1.5 is aligned.

\section{Related work}
\label{sec:related}

\paragraph{Particle decoding for language models.}
Sequential Monte Carlo has steered autoregressive language models through probabilistic
programs \citep{lew2023smc}, learned twists \citep{zhao2024twistedsmc}, syntactic and
semantic constraints \citep{loula2025smcconstraints}, and particle-based inference-time
reasoning \citep{puri2025reasoningsmc, markovic2026smc, giannone2025epf, giannone2026ipf,
tran2026pbsmc, nguyen2026apps}. Parallel work sharpens the base-model distribution or its
answer marginal rather than steering toward an external reward
\citep{karan2026powersampling, azizi2026powersmc, ji2026scalable, zhou2026cutting,
arzhantsev2026marginal, wang2023selfconsistency}. Feynman--Kac steering has also been
applied to image and text diffusion models using intermediate rewards
\citep{singhal2025general}. For diffusion LLMs, SR-SMC uses accumulated
decoder confidence \citep{luo2026srsmc}, $S^3$ uses a composite verifier \citep{s3smc2026},
POKE-SMC uses lookahead path likelihood \citep{liu2026pokesmc}, and TRI-TSMC learns
twisting functions \citep{tritsmc2026}. PG-DLM instead addresses one-shot weight
degeneracy by refining complete trajectories with particle Gibbs \citep{dang2026pgdlm}.
These methods modify the proposal, potential, target or outer refinement procedure; we
hold the guidance fixed and study lineage loss at a resampling event.

\paragraph{Resampling, genealogy and final-answer selection.}
Stratified, residual and systematic resampling reduce variance relative to multinomial
resampling and conventionally return equal weights \citep{kitagawa1996monte,
liu1998sequential, carpenter1999improved, douc2005comparison}.
\citet{fearnhead2003resampling} resample only the lighter particles, a thresholding
principle related to Chopthin's pass-through region \citep{gandy2016chopthin}, which
\citet{ahmadi2026ccps} applied to autoregressive power sampling; our lineage-based formulation, the input weight floor, and their integration with
block-wise diffusion SMC are new here. The loss of ancestral diversity under repeated resampling is
classical \citep{doucet2001sequential, delmoral2004feynman}; we make its mismatch with effective
sample size explicit in diffusion LLM decoding. Selecting a final answer is a separate problem,
addressed through voting or output clustering \citep{wang2023selfconsistency, chen2023usc,
kuhn2023semanticentropy, li2022alphacode, chen2023codet} or through verifiers and reward models
\citep{cobbe2021gsm8k, uesato2022process, lightman2024verify, zhang2025prmlessons,
zeng2025acecoder}; Section~\ref{sec:selectors} compares selection rules on a fixed population.

\section{Conclusion}
\label{sec:conclusion}

Multinomial resampling can remove useful lineages while resetting the effective
sample size, so high ESS does not imply genealogical diversity. LP-SMC addresses this
mismatch by combining Chopthin's unequal-weight resampling with an input weight floor.
Across two diffusion LLMs and four reasoning and
code-generation benchmarks it increases oracle coverage in every setting and accuracy under a common
reward-model selector in eleven of twelve, and the mechanism is visible in the populations
themselves: surviving lineages rise with the decoder left unchanged. Population management can
therefore widen the set of problems on which inference-time search produces a correct candidate; how
much of that becomes final accuracy depends on the selection rule; Appendix~\ref{app:limitations} states the limitations.

\section*{AI use statement}

Generative AI tools were used to assist with language editing, drafting and revising
portions of the manuscript, and creating or editing data-processing, evaluation, and plotting
scripts. Generative AI tools were also used for discussion and feedback during the research
process; all research decisions, including the final formulation of the method, experimental
design, mathematical claims, theoretical results, and interpretation of the results, were made and
verified by the authors. The theoretical results and proofs were derived and validated by the
authors. All AI-assisted text and code were reviewed and verified by the authors. Numerical results
reported in the paper were recomputed from the raw per-particle records. The authors take
responsibility for the final content of the paper, including AI-assisted text and code.

\section*{Reproducibility statement}

Section~\ref{sec:method} specifies the method: Chopthin resampling in
Section~\ref{sec:chopthin} and the input weight floor in Section~\ref{sec:cap}, with the floored
weights defined in \eqref{eq:cap} and the admissible range of $\kappa$ in \eqref{eq:kappastar}.
The Setup paragraph of Section~\ref{sec:experiments} lists every decoding hyperparameter
(generation length $L$, block size, decoding steps per block, commitment threshold, temperature,
particle count $N$, the ESS gate, the Chopthin ratio $\eta$ and the floor $\kappa$); notation is
collected in Appendix~\ref{app:notation}. Complete proofs are in Appendix~\ref{app:theory}:
Proposition~\ref{prop:essblind} in Appendix~\ref{app:essblind}, Lemma~\ref{lem:essfloor} in
Appendix~\ref{app:essfloor}, and the floor's survival bound in \eqref{eq:floorsurv} with its
mass-displacement statistics in Appendix~\ref{app:floor}. Appendix~\ref{app:prmscore} states the
external potential exactly, including prompt construction, step segmentation and score
aggregation \eqref{eq:prmscore}; Appendix~\ref{app:protocol} gives the evaluation protocol, with
shot counts, answer extraction and code grading; and Appendix~\ref{app:coderm} documents the code
reward models considered. Within each controlled comparison the model, guidance signal, evaluation
problems and decoding hyperparameters are fixed, with configurations differing only in the
population-management step. Every particle population is written to disk, so all reported
selectors are recomputed offline without regenerating samples, and the paired statistics of
Appendix~\ref{app:paired} are derived from those same records. Code, task definitions and
per-problem scores will be released.

\section*{Ethics statement}

This work studies inference-time decoding for open-weight language models. No human
subjects were involved and no new data were collected. Experiments use publicly available
benchmarks (MATH-500, GSM8K, HumanEval, and MBPP), publicly released model weights (LLaDA-1.5 and
Dream-7B), and publicly available reward models. The proposed method operates at decoding time and
does not train or modify the underlying model weights; consequently, generated outputs remain
subject to the limitations and failure modes of the underlying models and reward signals.
Evaluation on HumanEval and MBPP requires execution of model-generated code; this execution is
performed only through the evaluation harness's sandboxed execution path on compute controlled by
the authors.

\bibliography{iclr2027_conference,references_chopthin}
\bibliographystyle{iclr2027_conference}

\appendix
\clearpage
\section{Notation}
\label{app:notation}
The table below collects the symbols used in the paper, with the section or equation
that introduces each.
\vspace{-0.5\baselineskip}
\begin{center}
\footnotesize
\setlength{\tabcolsep}{4pt}
\begin{tabular}{lp{8.8cm}l}
\toprule
Symbol & Meaning & Introduced \\
\midrule
\multicolumn{3}{l}{\emph{Masked diffusion decoding}} \\
$L$ & generation length: the number of token positions in the sequence & Sec.~\ref{sec:decoding} \\
$\mathcal{V}$ & token vocabulary & Sec.~\ref{sec:decoding} \\
$t$, $T$ & decoding step, counted backward from $T$ to $1$; $T$ is the number of steps & Sec.~\ref{sec:decoding} \\
$x_t$ & partially masked sequence at step $t$; $x_T$ is fully masked and $x_0$ is the output & Sec.~\ref{sec:decoding} \\
$\mathcal{M}_t$ & masked positions of the current block that are eligible at step $t$ & Sec.~\ref{sec:decoding} \\
$p_\theta$ & the model's predictive distribution over tokens at a position, given $x_t$ & Sec.~\ref{sec:decoding} \\
$\hat x_0(j)$, $c_t(j)$ & token drawn at position $j$, and the confidence $p_\theta(\hat x_0(j) \mid x_t)$ of that draw & Sec.~\ref{sec:decoding} \\
$\mathcal{S}_t$ & commit set: the positions whose drawn tokens are committed at step $t$ & Sec.~\ref{sec:decoding} \\
\midrule
\multicolumn{3}{l}{\emph{Sequential Monte Carlo decoding}} \\
$N$ & number of particles & Sec.~\ref{sec:tilt} \\
$x^{(i)}_t$, $U^{(i)}_t$ & state and weight of particle $i$; the particle index is a superscript & Sec.~\ref{sec:tilt} \\
$K_t$ & transition kernel: the probability that one decoding step moves $x_t$ to $x_{t-1}$ & Sec.~\ref{sec:tilt} \\
$G_{t-1}$ & potential: the score assigned to the tokens committed at a step; in SR-SMC, the product of the committed confidences & Eq.~(\ref{eq:G}) \\
$U_t$ & unnormalized weight, $U_{t-1} = U_t\,G_{t-1}$ between resampling events & Eq.~(\ref{eq:sr-weight}) \\
$w_i$ & normalized weight of particle $i$, $U^{(i)} / \sum_k U^{(k)}$ & Sec.~\ref{sec:tilt} \\
\fixnew{$U_{\max}$, $w_{\max}$} & \fixnew{largest unnormalized and normalized weight in the population at a boundary} & \fixnew{Eq.~(\ref{eq:cap})} \\
\fixnew{$\tilde U^{(i)}$, $\tilde w_i$} & \fixnew{floored weight and normalized floored weight of particle $i$} & Eq.~(\ref{eq:cap}) \\
\fixnew{$s(x)$, $p$, $\beta$} & \fixnew{external potential: the log-odds $\beta \log\frac{p}{1-p}$ of the reward model's mean step score $p$ on the prefix, with $\beta = 1$; the weight is $U = \exp s(x)$} & \fixnew{Eq.~(\ref{eq:prmscore})} \\
$\mathrm{ESS}$ & effective sample size, $1 / \sum_i w_i^{2} \in [1, N]$ & Sec.~\ref{sec:tilt} \\
ESS gate & the test $\mathrm{ESS} < N/2$ at a block boundary; when it is met, resampling is triggered & Sec.~\ref{sec:tilt} \\
$\tau$ & a complete decoding path $(x_T, \dots, x_0)$, also called a trajectory & Sec.~\ref{sec:tilt} \\
$Q(\tau)$, $C(\tau)$ & decoder path distribution and accumulated potential; the target is proportional to $Q(\tau)\,C(\tau)$ & Sec.~\ref{sec:tilt} \\
\midrule
\multicolumn{3}{l}{\emph{Resampling operators}} \\
$A_k$ & ancestor of output particle $k$: the input particle whose state it continues & Sec.~\ref{sec:resamplers} \\
$w^{+}_k$ & weight returned for output particle $k$ & Sec.~\ref{sec:resamplers} \\
$O_i$ & number of offspring of input particle $i$ & Sec.~\ref{sec:resamplers} \\
$\eta$ & Chopthin's bound on the ratio of returned weights; $3 + \sqrt{8}$ throughout & Secs.~\ref{sec:resamplers}, \ref{sec:chopthin} \\
$a$ & Chopthin threshold, chosen so that the expected offspring counts sum to $N$ & Sec.~\ref{sec:resamplers} \\
$h^{\eta}_a(w_i)$ & expected number of offspring of particle $i$ & Eq.~(\ref{eq:h}) \\
\midrule
\multicolumn{3}{l}{\emph{Lineages}} \\
$r(i)$ & root of particle $i$: the label of the initial particle it descends from & Sec.~\ref{sec:ess-blind} \\
$R$ & number of surviving lineages, the number of distinct roots & Sec.~\ref{sec:ess-blind} \\
$\mathcal{P} = (w, r)$, $\mathcal{P}^{+}$ & the population before and after a resampling event & Sec.~\ref{sec:ess-blind} \\
$\rho$, $w_\rho$ & a root, and the total weight of its lineage & Prop.~\ref{prop:essblind} \\
\midrule
\multicolumn{3}{l}{\emph{Weight floor}} \\
$\kappa$ & weight floor: the largest gap, in nats, allowed between a weight and the maximum weight; $\ln N$ throughout & Eq.~(\ref{eq:cap}) \\
$\kappa^{\star}$ & critical floor at which the worst-case ESS lower bound equals $N/2$; the gate can fire only for $\kappa > \kappa^{\star}$, $\ln(3 + \sqrt{8})$ & Eq.~(\ref{eq:kappastar}) \\
\bottomrule
\end{tabular}
\end{center}

\section{Supporting results}
\label{app:theory}

\subsection{Chopthin's offspring draw, and the proof of Proposition~\ref{prop:essblind}}
\label{app:essblind}

The offspring function $h^{\eta}_a$ of~\eqref{eq:h}
(Section~\ref{sec:resamplers}) sets the expected number of offspring of particle $i$,
with $a$ chosen so that the expected offspring counts sum to $N$, $\sum_{i=1}^{N} h^{\eta}_a(w_i) = N$. Chopthin then couples the integer parts of these counts with randomized fractional remainders so that the realized population contains exactly $N$ particles, and a weight correction proportional to those remainders preserves the total weight \citep{gandy2016chopthin, ahmadi2026ccps}.

\begin{proof}[Proof of Proposition~\ref{prop:essblind}]
(i) After the reset every weight equals $1/N$, so $\mathrm{ESS} = N$. Lineage $\rho$
survives if and only if at least one of the $N$ independent ancestor draws falls in $\rho$,
which has probability $1 - (1 - w_\rho)^{N}$; summing over lineages gives the expectation.
For uniform weights put $w_i = 1/N$; the sequence $(1 - 1/N)^{N}$ increases to
$e^{-1}$.
(ii) As a function of $a$ the total $\sum_i h^{\eta}_a(w_i)$ is continuous and never
increases. It reaches $N$ with every term equal to one precisely when $2 w_i / \eta \le a \le w_i$ for all $i$,
that is, when $a \in [\,2 w_{\max} / \eta,\; w_{\min}\,]$, writing $w_{\max} = \max_i w_i$ and $w_{\min} = \min_i w_i$, an interval that is non-empty
if and only if $w_{\max} / w_{\min} \le \eta / 2$.

Under the hypothesis $w_{\max} / w_{\min} \le \eta / 2$, these are the only solutions. For
$a > w_{\min}$ some term is below one and none is above one; for $a < 2 w_{\max} / \eta$ some
term is above one and none is below one.

For any such $a$ every particle has expected offspring count exactly one with zero
fractional part, so it receives exactly one offspring. No particle is thinned, the
correction term is zero, and each particle returns with its own weight. Hence $\mathcal{P}^{+} =
\mathcal{P}$.
\end{proof}

\subsection{The ESS lower bound implied by a weight-ratio bound}
\label{app:essfloor}

\begin{lemma}[ESS lower bound]
\label{lem:essfloor}
Let $w_1, \dots, w_N > 0$ with $\max_i w_i / \min_i w_i \le \eta$ for some $\eta \ge 1$.
Then
\[
  \mathrm{ESS}(w) = \frac{(\sum_i w_i)^{2}}{\sum_i w_i^{2}} \;\ge\; \frac{4 \eta N}{(\eta + 1)^{2}} .
\]
At $\eta = 3 + \sqrt{8}$ the right-hand side equals $N / 2$.
\end{lemma}

\begin{proof}
The ESS does not change when all weights are multiplied by the same constant, so assume
$\min_i w_i = 1$; then $1 \le w_i \le \eta$ for every $i$. Hence $(w_i - 1)(\eta - w_i)
\ge 0$, that is, $w_i^{2} \le (\eta + 1) w_i - \eta$. Summing over $i$ and writing
$s = \sum_i w_i \in [N, \eta N]$ gives $\sum_i w_i^{2} \le (\eta + 1) s - \eta N$, and
therefore
\[
  \mathrm{ESS}(w) \;\ge\; g(s) := \frac{s^{2}}{(\eta + 1) s - \eta N} .
\]
The denominator is positive on $[N, \eta N]$. The derivative of $g$ has the sign of
$(\eta + 1) s - 2 \eta N$, so $g$ is smallest at $s = 2 \eta N / (\eta + 1)$, which lies
in $[N, \eta N]$, and there $g = 4 \eta N / (\eta + 1)^{2}$. For $\eta = 3 + \sqrt{8}$,
$4\eta / (\eta + 1)^{2} = 1/2$.
\end{proof}

\citet[Lemma~2]{gandy2016chopthin} state the weaker bound $4(\eta N + 1 - \eta^{2}) /
(\eta + 1)^{2}$, which has the same leading term; the difference is the constant
$4(\eta^{2} - 1)/(\eta + 1)^{2} = 4(\eta - 1)/(\eta + 1)$, equal to $2\sqrt{2}$ at
$\eta = 3 + \sqrt{8}$. The
bound concerns the weights only; it says nothing about the number of surviving lineages.

\fixnew{\subsection{The weight floor: multinomial survival bound and mass displacement}
\label{app:floor}
% Moved here from Section~\ref{sec:cap} on 2026-09-25 (page budget): the two weight
% profiles with eq:floorsurv and the mass-displacement statistics. Equation eq:floorsurv is
% unchanged. The gate-compatibility paragraph was removed later the same day (reviewer).

\paragraph{Multinomial survival bound.}
In the ESS-minimising admissible profile, one particle at the maximum and $N - 1$ at the
floor, each floored particle has normalized weight $1/(e^{\kappa} + N - 1)$, that is
$1/(2N-1)$ at $\kappa = \ln N$, so a multinomial ancestor draw keeps it with probability
$1 - (1 - 1/(2N-1))^{N}$, $0.41$ at $N = 16$. The profile that minimises a floored weight
is different: the floored weight is smallest when the other $N-1$ particles sit at the
maximum, which gives $1/(N(N-1)+1)$, $1/241$ at $N = 16$. Under a multinomial ancestor
draw the floor therefore gives
\begin{equation}
  \Pr[\text{survive}] \;\ge\; 1 - \left(1 - \frac{1}{N(N-1)+1}\right)^{\!N},
  \qquad 0.064 \text{ at } N = 16 .
  \label{eq:floorsurv}
\end{equation}

\paragraph{Mass displacement.}
The mass moved into the tail at one boundary is at most $(N-1)/(2N-1)$ of the total,
$0.48$ at $N = 16$, attained only when $N-1$ particles have weight equal to the floor. Across the
$10{,}552$ \fixnew{block-boundary observations} of LLaDA-1.5 on GSM8K \fixnew{($1{,}319$
problems $\times$ $8$ blocks; the ESS gate fired at $2{,}963$ of them)}, $23.8\%$ of particle slots have weight equal to the floor. Floored slots hold $0.057$ of the normalized mass per \fixnew{boundary} on average, which
upper-bounds the mass moved by the floor; the same statistic is $0.002$ on the unfloored
configurations (SR-SMC and Chopthin), where it only counts weights that fall near
$w_{\max} e^{-\kappa}$ by chance. This upper bound is about an order of magnitude below
the worst case.}

% Figure moved here from Section~\ref{sec:ablation} on 2026-09-25 (page budget); the
% caption is self-contained.
% Figure: floor-depth sweep (generated by make_fig_floor_sweep.py; edit the script,
% not this file). Needs \usepackage{tikz} only -- same requirement as fig_oracle.tex.
\begin{figure}[!ht]
\centering
\definecolor{fsAcc}{RGB}{40,74,160}
\definecolor{fsLin}{RGB}{14,122,120}
\begin{tikzpicture}[x=1cm,y=1cm]
\tikzset{
  ax/.style={draw=black!55, line width=0.4pt},
  gl/.style={draw=black!12, line width=0.3pt},
  tk/.style={font=\fontsize{6.0}{6.8}\selectfont, text=black!70},
  lb/.style={font=\fontsize{6.4}{7.2}\selectfont},
  vl/.style={font=\fontsize{6.0}{6.8}\selectfont},
  ser/.style={line width=0.7pt},
}
\draw[gl] (1.000,1.259) -- (13.200,1.259);
\node[tk, anchor=east] at (0.940,1.259) {49};
\draw[gl] (1.000,1.816) -- (13.200,1.816);
\node[tk, anchor=east] at (0.940,1.816) {51};
\draw[gl] (1.000,2.373) -- (13.200,2.373);
\node[tk, anchor=east] at (0.940,2.373) {53};
\draw[gl] (1.000,2.930) -- (13.200,2.930);
\node[tk, anchor=east] at (0.940,2.930) {55};
\draw[gl] (1.000,0.000) -- (13.200,0.000);
\node[tk, anchor=east] at (0.940,0.000) {0};
\draw[gl] (1.000,0.300) -- (13.200,0.300);
\node[tk, anchor=east] at (0.940,0.300) {8};
\draw[gl] (1.000,0.600) -- (13.200,0.600);
\node[tk, anchor=east] at (0.940,0.600) {16};
\draw[ax] (1.000,0.980) -- (1.000,2.930);
\draw[ax] (1.000,0.000) -- (1.000,0.600);
\draw[ser, fsAcc] (2.350,1.259) -- (4.725,1.537) -- (7.100,2.039) -- (9.475,2.206) -- (11.850,2.540);
\fill[fsAcc] (2.350,1.259) circle (0.060cm);
\node[vl, text=fsAcc, anchor=south] at (2.350,1.359) {49.0};
\fill[fsAcc] (4.725,1.537) circle (0.060cm);
\node[vl, text=fsAcc, anchor=south] at (4.725,1.637) {50.0};
\fill[fsAcc] (7.100,2.039) circle (0.060cm);
\node[vl, text=fsAcc, anchor=south] at (7.100,2.139) {51.8};
\fill[fsAcc] (9.475,2.206) circle (0.060cm);
\node[vl, text=fsAcc, anchor=south] at (9.475,2.306) {52.4};
\fill[fsAcc] (11.850,2.540) circle (0.060cm);
\node[vl, text=fsAcc, anchor=south] at (11.850,2.640) {53.6};
\draw[ser, fsLin, dash pattern=on 2pt off 1.4pt] (2.350,0.102) -- (4.725,0.147) -- (7.100,0.198) -- (9.475,0.262) -- (11.850,0.437);
\fill[white] (2.350,0.102) circle (0.060cm);
\draw[fsLin, line width=0.55pt] (2.350,0.102) circle (0.060cm);
\node[vl, text=fsLin, anchor=south] at (2.350,0.192) {2.71};
\fill[white] (4.725,0.147) circle (0.060cm);
\draw[fsLin, line width=0.55pt] (4.725,0.147) circle (0.060cm);
\node[vl, text=fsLin, anchor=south] at (4.725,0.237) {3.93};
\fill[white] (7.100,0.198) circle (0.060cm);
\draw[fsLin, line width=0.55pt] (7.100,0.198) circle (0.060cm);
\node[vl, text=fsLin, anchor=south] at (7.100,0.288) {5.29};
\fill[white] (9.475,0.262) circle (0.060cm);
\draw[fsLin, line width=0.55pt] (9.475,0.262) circle (0.060cm);
\node[vl, text=fsLin, anchor=south] at (9.475,0.352) {6.98};
\fill[white] (11.850,0.437) circle (0.060cm);
\draw[fsLin, line width=0.55pt] (11.850,0.437) circle (0.060cm);
\node[vl, text=fsLin, anchor=south] at (11.850,0.527) {11.65};
\node[lb, anchor=north] at (2.350,-0.130) {Multinomial};
\node[tk, anchor=north] at (2.350,-0.430) {none};
\node[lb, anchor=north] at (4.725,-0.130) {Chopthin};
\node[tk, anchor=north] at (4.725,-0.430) {none};
\node[lb, anchor=north] at (7.100,-0.130) {Chopthin + floor};
\node[tk, anchor=north] at (7.100,-0.430) {$1/32$};
\node[lb, anchor=north] at (9.475,-0.130) {Chopthin + floor};
\node[tk, anchor=north] at (9.475,-0.430) {$1/16$};
\node[lb, anchor=north] at (11.850,-0.130) {Chopthin + floor};
\node[tk, anchor=north] at (11.850,-0.430) {$1/8$};
\node[tk, anchor=north] at (7.100,-0.720) {floor as a fraction of the maximum weight};
\node[lb, rotate=90, anchor=south] at (0.540,1.955) {Accuracy (\%)};
\node[lb, rotate=90, anchor=south] at (0.540,0.300) {Lineages};
\fill[fsAcc] (2.700,3.310) circle (0.060cm);
\node[tk, anchor=west] at (2.820,3.310) {reward-model accuracy};
\fill[white] (6.600,3.310) circle (0.060cm);
\draw[fsLin, line width=0.55pt] (6.600,3.310) circle (0.060cm);
\node[tk, anchor=west] at (6.720,3.310) {surviving lineages (of $16$)};
\end{tikzpicture}
\caption{Floor depth on MATH-500 (LLaDA-1.5, $N = 16$, reward-model selector). From
left to right: multinomial resampling, Chopthin without the floor, and Chopthin with the
floor at $1/32$, $1/16$ ($\kappa = \ln N$, used throughout) and $1/8$ of the maximum
weight. Among the Chopthin configurations, accuracy and surviving lineages rise monotonically as
the floor is raised.}
\label{fig:floorsweep}
\end{figure}
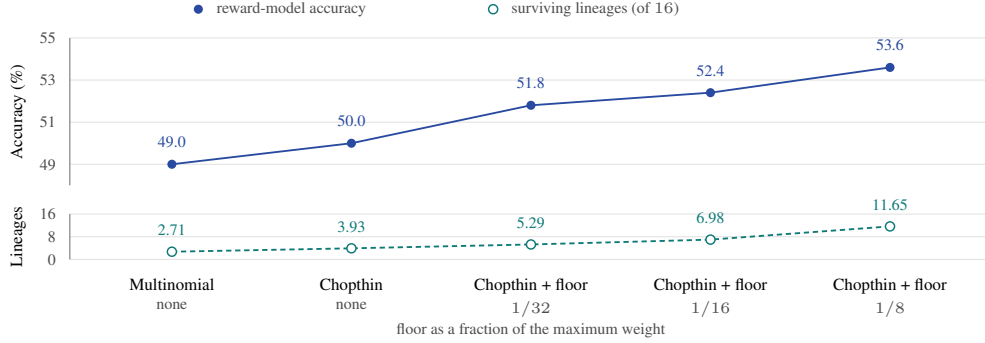

\fixnew{\subsection{Selection rules by model}
\label{app:selectors}
The four selection rules of Section~\ref{sec:selectors} are: \emph{confidence}, the argmax of the
decoder's cumulative confidence and the rule of the released SR-SMC code; \emph{majority}, a vote
over all $N$ particles; \emph{distinct}, a vote over distinct outputs merged by the grader's answer
equivalence (\citealp{ahmadi2026ccps}; conventions in Appendix~\ref{app:protocol}); and
\emph{reward model}, the argmax of Qwen2.5-Math-PRM-7B on mathematics or AceCodeRM-7B on code,
applied identically to both populations. Table~\ref{tab:selectors} reports all four on LLaDA-1.5;
Table~\ref{tab:dreamselfull} reports the same for Dream-7B.}
% Table moved here from Section~\ref{sec:selectors} on 2026-09-25 (page budget).
% [!ht] rather than [t]: with [t] the table floated above its own B.4 heading.
\begin{table}[!ht]
\caption[Accuracy under four selection rules applied to the same final populations
(LLaDA-1.5).]{Accuracy under four selection rules applied to the same final populations
(LLaDA-1.5, $N{=}16$, $L{=}256$). \fixcut{The reward model enters in two independent
places and the table separates them. \emph{Guidance} is where it acts during generation:
Internal uses the decoder's confidence in the resampling weight, External PRM puts
Qwen2.5-Math-PRM-7B in the weight. The \emph{Reward model} column is a selection rule
applied afterwards to a finished population, argmax of Qwen2.5-Math-PRM-7B (mathematics)
or AceCodeRM-7B (code); it is available under either guidance. The other selectors read
only the population: confidence, argmax of the decoder's cumulative confidence and the
rule of the released SR-SMC code; majority, vote over all particles; distinct, vote over
distinct outputs merged by answer equivalence.}\fixnew{\emph{Guidance} is where the reward
model acts during generation; the \emph{Reward model} column is a selection rule applied
afterwards to the finished population (Qwen2.5-Math-PRM-7B on mathematics, AceCodeRM-7B on
code) and is available under either guidance.} Bold: better within a pair.}
\label{tab:selectors}
\begin{center}
\footnotesize
\setlength{\tabcolsep}{3.5pt}
\begin{tabular}{lllcccc}
\toprule
Benchmark & Guidance & Population management & Confidence & Majority & Distinct & Reward model \\
\midrule
\multirow{4}{*}{\shortstack[l]{\textbf{MATH-500}\\$n = 500$}}
 & \multirow{2}{*}{Internal}     & SR-SMC          & \textbf{41.8} & 41.4 & 41.2 & 49.0 \\
 &                               & LP-SMC          & 39.2 & \textbf{44.4} & \textbf{45.0} & \textbf{52.4} \\
\cmidrule(lr){2-7}
 & \multirow{2}{*}{External PRM} & Multinomial SMC & \textbf{48.8} & \textbf{52.6} & \textbf{51.4} & 53.8 \\
 &                               & LP-SMC          & 43.6 & \textbf{52.6} & 50.2 & \textbf{55.0} \\
\midrule
\multirow{4}{*}{\shortstack[l]{\textbf{GSM8K}\\$n = 1319$}}
 & \multirow{2}{*}{Internal}     & SR-SMC          & 82.7 & 84.4 & 84.8 & 88.2 \\
 &                               & LP-SMC          & \textbf{83.5} & \textbf{86.4} & \textbf{86.8} & \textbf{90.5} \\
\cmidrule(lr){2-7}
 & \multirow{2}{*}{External PRM} & Multinomial SMC & \textbf{88.0} & 91.4 & 91.1 & 90.4 \\
 &                               & LP-SMC          & 85.6 & \textbf{92.3} & \textbf{91.2} & \textbf{90.8} \\
\midrule
\multirow{2}{*}{\shortstack[l]{\textbf{HumanEval}\\$n = 164$}}
 & \multirow{2}{*}{Internal}     & SR-SMC          & \textbf{43.3} & \textbf{41.5} & 40.9 & 48.2 \\
 &                               & LP-SMC          & 41.5 & \textbf{41.5} & \textbf{41.5} & \textbf{56.1} \\
\midrule
\multirow{2}{*}{\shortstack[l]{\textbf{MBPP}\\$n = 482$}}
 & \multirow{2}{*}{Internal}     & SR-SMC          & 40.5 & 40.9 & \textbf{41.3} & 49.8 \\
 &                               & LP-SMC          & \textbf{40.9} & \textbf{41.3} & \textbf{41.3} & \textbf{50.6} \\
\bottomrule
\end{tabular}
\end{center}
\end{table}

\label{app:dreamsel}

Two conventions differ from Table~\ref{tab:selectors} and are stated here. First, $40.4\%$ of Dream's particles on MATH-500 finish without a
\texttt{\textbackslash boxed} expression, against $3.5\%$ for LLaDA; empty answers are
excluded from the vote in both tables, but the exclusion removes far more of the population
here. Second, answer equivalence on code is exact program identity after AST
canonicalisation, as in Table~\ref{tab:selectors}, except that the key is the sanitised
program the grader executes rather than the raw generation. Dream continues into the next
few-shot example after finishing, so its raw text does not parse and the raw key would
measure the parser, not the selector. The code reward model, AceCodeRM-7B, is scored on the
sanitised program for Dream for the same reason.

\begin{table}[!ht]
\caption[Dream-7B under every selection rule.]{Dream-7B under every selection rule.
As in Table~\ref{tab:selectors}, \emph{Guidance} is
where the reward model acts during generation and the \emph{Reward model} column is a
selection rule applied afterwards to a finished population; the two are independent.
\fixnew{Bold: better within a pair.}
\fixcut{Coverage: Table 2; the MATH-500 coverage pairs, which it does not carry, are
44.2 / 58.8 under Internal guidance and 59.0 / 62.0 under External PRM, and on GSM8K under
External PRM 92.3 / 95.0.}}
\label{tab:dreamselfull}
\centering
\footnotesize
\setlength{\tabcolsep}{3.5pt}
\begin{tabular}{lllcccc}
\toprule
Benchmark & Guidance & Population management & Confidence & Majority & Distinct & Reward model \\
\midrule
\textbf{MATH-500} & \multirow{2}{*}{Internal}     & SR-SMC          & \textbf{37.8} & 42.2 & 42.6 & 41.8 \\
$n = 500$         &                               & LP-SMC          & 33.4 & \textbf{47.0} & \textbf{49.6} & \textbf{46.8} \\
\cmidrule(lr){2-7}
                  & \multirow{2}{*}{External PRM} & Multinomial SMC & \textbf{44.0} & 54.2 & 53.6 & \textbf{49.2} \\
                  &                               & LP-SMC          & 40.8 & \textbf{55.0} & \textbf{54.6} & 48.2 \\
\midrule
\textbf{GSM8K}    & \multirow{2}{*}{Internal}     & SR-SMC          & \textbf{77.6} & \textbf{77.7} & 78.1 & 78.2 \\
$n = 1319$        &                               & LP-SMC          & 70.6 & 72.9 & \textbf{81.3} & \textbf{86.7} \\
\cmidrule(lr){2-7}
                  & \multirow{2}{*}{External PRM} & Multinomial SMC & \textbf{87.2} & 89.1 & \textbf{89.0} & 90.8 \\
                  &                               & LP-SMC          & 80.6 & \textbf{89.3} & 88.1 & \textbf{91.7} \\
\midrule
\textbf{HumanEval}& \multirow{2}{*}{Internal}     & SR-SMC          & 46.3 & 45.7 & 47.0 & 48.8 \\
$n = 164$         &                               & LP-SMC          & \textbf{48.2} & \textbf{47.0} & \textbf{47.6} & \textbf{67.1} \\
\midrule
\textbf{MBPP}     & \multirow{2}{*}{Internal}     & SR-SMC          & \textbf{60.8} & \textbf{60.2} & \textbf{59.8} & 59.8 \\
$n = 482$         &                               & LP-SMC          & 58.1 & 55.6 & 56.6 & \textbf{62.0} \\
\bottomrule
\end{tabular}
\end{table}

% Subsection "Separating the resampler from the floor: protocol" (label app:capsep, unused)
% removed 2026-09-25 (reviewer). Heading reproduced as bold text inside \fixcut because
% \sout cannot hold \subsection; cross-references in plain text.
\fixcut{\textbf{Separating the resampler from the floor: protocol.} Table 3 supports the
comparison in Section 5.2. Every configuration uses the decoder's own confidence as the potential,
the same ESS${<}N/2$ gate, the same seeds and the same $1319$ GSM8K documents. The
multinomial and Chopthin rows without the floor repeat the GSM8K line of Table 2 so that
every configuration can be read against the others in one place. Systematic resampling is reported
both without and with the floor, so the six rows form a complete factorial of ancestor
draw against floor and each factor can be read at fixed levels of the other. The coverage
coincidence in the two unfloored rows is exact:
multinomial and systematic resampling each solve $1198$ of the $1319$ problems,
overlapping on $1155$, with $43$ solved by one configuration and not the other in each direction.}

\subsection{The external potential, exactly}
\label{app:prmscore}

\fixnew{Section~\ref{sec:experiments} writes the externally guided weight as
$U^{(i)} = \exp s(x^{(i)})$. This section states $s$ exactly, since the reward model's raw
output is a per-step probability, not a log-weight.

At a block boundary the committed prefix of particle $i$ is decoded to text and split into
steps on blank lines; empty steps are dropped, and a prefix with no non-empty step is
assigned $p = 0.5$. A chat sequence is built with the system prompt \emph{``Please reason
step by step, and put your final answer within \texttt{\textbackslash boxed\{\}}.''}, the
question as the user turn, and the steps joined by the separator token
\texttt{<extra\_0>} with a trailing separator as the assistant turn. The question is the
final few-shot example only: the prompt is split on the last occurrence of its family's
start marker, so the in-context examples are not scored. The sequence is truncated to its
first $4096$ tokens.

Qwen2.5-Math-PRM-7B emits a two-way distribution at every \texttt{<extra\_0>} position. Let
$p_k$ be the positive-class probability at step $k$ and $p = \operatorname{mean}_k p_k$,
clamped to $[10^{-3}, 1-10^{-3}]$. \fixnew{The index $k$ runs over all steps of the prefix,
that is over every \texttt{<extra\_0>} position surviving the $4096$-token truncation, so
the mean is unweighted over the whole committed prefix, not the score of the final step alone.} Then
\begin{equation}
  s(x) \;=\; \beta \, \log \frac{p}{1-p}, \qquad \beta = 1 ,
  \label{eq:prmscore}
\end{equation}
so $s$ is a log-odds and $U = \exp s = (p/(1-p))^{\beta}$. We use the \emph{absolute} form:
at each boundary $\log w$ is replaced by $s$ instead of incremented by it, which is the
boundary-wise reweighting noted in Section~\ref{sec:experiments}. \fixcut{Steps are not
capped; an earlier version scored only the first $20$ steps, which truncated the final
\texttt{\textbackslash boxed\{\}} of long answers, and the $4096$-token limit is the only
bound in the reported runs.}}

\fixnew{\subsection{Evaluation protocol}
\label{app:protocol}
% Moved here from the Setup paragraph of Section~\ref{sec:experiments} and from
% Section~\ref{sec:selectors} on 2026-09-25 (page budget).
The shot counts of Section~\ref{sec:experiments} are the harness defaults for each
benchmark; they are identical across configurations and were not tuned. Mathematics answers are
taken from the last boxed expression and compared by symbolic equivalence; code is graded
by executing the candidate, truncated at the task's stop marker and then reduced to its
longest parsable prefix, against the benchmark tests. A path $\tau$, its output $x_0$ and
the answer the grader extracts are distinct objects: several paths can give one output and
several outputs one answer, and distinct-output counts are counts of $x_0$. For
distinct-output consensus, the $N$ outputs of the last block are merged into their
distinct outputs, which are grouped into answers by the grader's equivalence, numeric or
symbolic on mathematics and exact program identity after AST canonicalisation on code,
with benchmark execution used only for final grading; the answer held by the largest
number of distinct outputs is returned \citep{ahmadi2026ccps}.
Appendix~\ref{app:dreamsel} states the conventions specific to Dream-7B.

The population-management operations introduce no direct decoder or reward-model
evaluations. Measured end to end on GSM8K with LLaDA-1.5, the mean number of network
evaluations per problem is $701.6$ for SR-SMC, $692.1$ for Chopthin and $708.0$ for
LP-SMC, and the mean wall time is $120.8$, $121.1$ and $119.6$ seconds on the same
hardware.}

\section{Reward models for code generation}
\label{app:coderm}

The reward model used on MATH-500 and GSM8K, Qwen2.5-Math-PRM-7B, is trained on
mathematical reasoning and is not validated for Python. This appendix records what is available
instead, since the availability of a step-level code reward model determined which
experiments were feasible.
\fixnew{We did not audit either reward model for overlap with the evaluation sets, and neither
model card documents its training corpus at a level that would settle the question. Overlap
would affect the absolute selected accuracies\fixcut{; it does not affect the comparison
between configurations, since the same selector is applied to identically paired populations}.}

\fixnew{\paragraph{No identified step-level code reward model meets all requirements.}
Most of the open-weight code reward models we identified are outcome models: they are
trained to score completed programs, so they can rerank a final population but provide no
prefix-level signal for the block-boundary resampling weights used here. Step-level code
reward models do exist, and we identified two, but neither simultaneously satisfies the
three requirements a usable one must meet here: released weights, a declared license, and
step-level scoring demonstrated on function completion.} CodePRM \citep{li2025codeprm} ships two LoRA adapters
in a source repository with no declared license, and SWE-Shepherd \citep{sweshepherd2026}
releases an adapter whose model card declares no license and which targets repository-agent
actions on SWE-Bench. Our decision rests on licensing and task fit, not on training-distribution mismatch: the runs
reported below use a mathematics process reward model on Python, which is further from its
training distribution than a code model evaluated on other code benchmarks would be. Of the outcome models we use
AceCodeRM-7B \citep{zeng2025acecoder} to rank finished programs (Table~\ref{tab:main}).

\paragraph{Execution as a reward, and why it is not applicable here.} A reward can be
obtained by running the candidate instead of scoring it with a model, and
\citet{markovic2026smc} do exactly this for HumanEval, combining unit-test execution with
an auxiliary syntax term. Execution needs a complete program, which they obtain by rolling
each particle forward to completion before scoring it. That is a change of method rather
than of reward: our sampler adds no model evaluations beyond decoding
(Section~\ref{sec:chopthin}), and lookahead rollouts would multiply that cost by the
remaining sequence length. The tests would also have to be separated from the evaluator,
which is the same difficulty as selecting on the visible tests below. A syntax-only reward
avoids both problems and is computable on an unfinished prefix, but it scores well-formedness, not correctness, so we do not report it.

\begin{table}[ht]
\caption{\fixnew{Open-weight code reward models considered for function-completion
experiments (September 2026).} Type O: outcome model, one score per finished program;
P: step-level process model, the only kind we regard as suitable inside a resampling weight.
\fixnew{SWE-Shepherd is not listed because it scores repository-agent actions on SWE-Bench
rather than function-completion steps.}}
\label{tab:coderm}
\begin{center}
\footnotesize
\setlength{\tabcolsep}{3.5pt}
\begin{tabular}{lllc}
\toprule
Model & Repository & License & Type \\
\midrule
AceCodeRM-7B/32B & \texttt{TIGER-Lab/AceCodeRM-*} & MIT & O \\
CodeScaler-1.7B--8B & \texttt{LARK-Lab/CodeScaler-*} & MIT & O \\
Themis-RM-0.6B--32B & \texttt{project-themis/Themis-RM-*} & Apache-2.0 & O \\
CodePRM & \texttt{SIMONLQY/CodePRM} & not declared & P \\
\bottomrule
\end{tabular}
\end{center}
\end{table}

\paragraph{The resampler under a task-mismatched reward.} The absence of a
usable code reward model leaves open the question: does the resampler's effect on the
population depend on the reward being well matched to the task? We can test that directly,
because a mathematics process reward model can be placed in the weight on code even though
it is strongly out of distribution there. Table~\ref{tab:oodprm} reports both configurations on HumanEval and MBPP
with Qwen2.5-Math-PRM-7B in the weight, on the same problems\fixcut{ and seeds}, so the two configurations
\fixnew{differ only in the population step (Chopthin resampling with the floor)}. Coverage rises in both benchmarks, by $4.3$ points on
HumanEval and $2.3$ on MBPP\fixcut{, so the gain does not require the reward to be
informative about the task}. \fixnew{Accuracy under confidence selection falls in both out-of-distribution guided code
experiments.} These rows are reported here and not in Table~\ref{tab:main} because the reward is not a code reward model, and describing
it as external code guidance would overstate the experiment.

\begin{table}[ht]
\caption{Both configurations on code under External PRM guidance, the mathematics PRM in the
weight and so out of distribution by construction. Guidance is identical across configurations,
so the comparison isolates the LP-SMC population-management intervention.}
\label{tab:oodprm}
\begin{center}
\footnotesize
\setlength{\tabcolsep}{5pt}
\begin{tabular}{llcc}
\toprule
Benchmark & Population management & Confidence & Coverage \\
\midrule
\multirow{2}{*}{HumanEval ($n{=}164$)}
 & Multinomial SMC           & \textbf{50.0} & 59.1 \\
 & LP-SMC                    & 42.7 & \textbf{63.4} \\
\midrule
\multirow{2}{*}{MBPP ($n{=}482$)}
 & Multinomial SMC           & \textbf{42.7} & 58.1 \\
 & LP-SMC                    & 41.5 & \textbf{60.4} \\
\bottomrule
\end{tabular}
\end{center}
\end{table}

\paragraph{Why we do not select with the visible tests.} Running each candidate against the
tests shown in the prompt would cost less than a reward model, but in MBPP \citep{austin2021mbpp} the three assertions shown in the prompt are the test cases used for grading, and in HumanEval \citep{chen2021humaneval} the docstring examples recur among the assertions of the hidden \texttt{check} function, so selecting on them would reuse information from the benchmark's own evaluator and break the separation between selection and evaluation. Rerankers that execute model-generated tests instead of the benchmark's own tests avoid this by construction \citep{ma2025coderm}.

% \clearpage removed 2026-09-25 (let the appendix flow)
\section{Lineage deletion risk}
\label{app:lineagerisk}

For each problem the ancestor lists recorded at every block are replayed forward to
recover which founding lineage each particle slot belongs to at each resampling event. The
final answers are graded, the lineages that produced a correct answer are identified, and
for each resampling event the total normalized weight of such a lineage, $w_\rho$, is used
to evaluate the probability that multinomial resampling would have left it without
offspring, $(1 - w_\rho)^{N}$. \fixnew{The higher median risk under LP-SMC is expected: LP-SMC admits correct lineages whose
intermediate weights are precisely those multinomial resampling would have been likely to remove.}
\fixnew{Table~\ref{tab:lineagerisk} reports, for each problem, the largest such value over
the joint range of both indices: over every boundary at which the ESS gate actually fired,
and over every lineage that produced a correct answer. Where several correct lineages exist
they are neither pooled nor selected between, so the statistic describes the most endangered
correct lineage at its most dangerous event rather than the fate of the problem's coverage.
Weights are pooled only within a lineage. Only problems with at least one correct final
answer enter the statistic, so the denominator differs between configurations.}
\fixnew{Under LP-SMC the correct lineage faced better-than-even odds of deletion on
$95.7\%$ of the problems Dream answers correctly and $87.4\%$ of those LLaDA answers
correctly, with median worst-case risk $0.79$ and $0.78$. No correct lineage in either
floored configuration faced a deletion probability above $0.9$, compared with $2.7\%$ and $9.5\%$ of
them under multinomial resampling \fixnew{and $6.2\%$ and $23.4\%$ under Chopthin without the
floor}: flooring a weight at $1/N$ of the maximum weight bounds $(1 - w_\rho)^{N}$
away from one. \fixnew{The controlled contrast is the last pair: holding the ancestor draw
fixed at Chopthin, adding the floor takes the fraction above $0.9$ from $6.2\%$ to zero on
Dream-7B and from $23.4\%$ to zero on LLaDA-1.5, so this is the floor's effect and not the
resampler's. Chopthin alone raises the fraction above the multinomial baseline.} These percentages are taken over different denominators (column Correct of
Table~\ref{tab:lineagerisk}), and each configuration is measured on its own weights, not as a paired counterfactual.}
% Moved here from the former Section 5.4 on 2026-09-25 (page budget).

\begin{table}[h]
\caption{Worst-case deletion probability $(1-w_\rho)^N$ faced by \fixnew{a} lineage
that produced a correct answer, GSM8K ($n{=}1319$): median, and the fraction of
correctly answered problems above each threshold. \fixnew{Per problem the maximum is taken
jointly over every fired resampling event and every correct lineage.}}
\label{tab:lineagerisk}
\begin{center}
\footnotesize
\setlength{\tabcolsep}{5pt}
\begin{tabular}{llccccc}
\toprule
Model & Population management & Correct & Median & $>0.5$ & $>0.8$ & $>0.9$ \\
\midrule
\multirow{2}{*}{LLaDA-1.5}
 & SR-SMC                & 1198 & 0.625 & 64.4\% & 24.6\% & 9.5\% \\
 & LP-SMC                & 1243 & 0.780 & 87.4\% & 31.1\% & \textbf{0.0\%} \\
\midrule
\multirow{2}{*}{Dream-7B}
 & SR-SMC                & 1036 & 0.118 & 20.5\% & 6.4\%  & 2.7\% \\
 & LP-SMC                & 1186 & 0.792 & 95.7\% & 42.1\% & \textbf{0.0\%} \\
\bottomrule
\end{tabular}
\end{center}
\end{table}

% \clearpage removed 2026-09-25 (let the appendix flow)
\clearpage
\section{Paired statistics for Table~\ref{tab:main} and Figure~\ref{fig:oracle}}
\label{app:paired}

Every comparison in Table~\ref{tab:main} and Figure~\ref{fig:oracle} is paired: \fixcut{a problem
receives the same seed under both configurations, so }the two configurations are evaluated on identical problems.
This appendix reports, for each pair, the discordant counts $a$--$b$ ($a$ = problems LP-SMC
solves and the baseline does not, $b$ = the reverse; concordant problems carry no information
about the difference) a $95\%$ confidence interval on the difference from $10{,}000$ bootstrap resamples over
problems, and the two-sided exact McNemar $p$-value computed from the same counts. Accuracy is
under the common reward-model selector; coverage is the quantity plotted in
Figure~\ref{fig:oracle}.

\fixnew{Every interval that excludes zero does so in LP-SMC's favour: eleven of the twelve
coverage comparisons and six of the twelve accuracy comparisons. None of the twenty-four
comparisons favours the baseline.}

\begin{table}[ht]
\caption{Paired statistics for every cell of Table~\ref{tab:main} and every pair of
Figure~\ref{fig:oracle}. $a$--$b$: discordant counts.
CI: percentile bootstrap on the difference, $10{,}000$ resamples over problems.
\fixnew{Differences are computed from the underlying counts; displayed accuracy and coverage
percentages elsewhere are rounded.}}
\label{tab:paired}
\begin{center}
\footnotesize
\setlength{\tabcolsep}{3pt}
\begin{tabular}{llllrrrlr}
\toprule
Model & Guidance & Benchmark & Quantity & $n$ & $a$--$b$ & $\Delta$ & $95\%$ CI & $p$ \\
\midrule
LLaDA-1.5 & Internal & MATH-500 & Accuracy & $500$ & $44$--$27$ & $+3.4$ & $[+0.2, +6.8]$ & 0.057 \\
LLaDA-1.5 & Internal & GSM8K & Accuracy & $1319$ & $76$--$45$ & $+2.4$ & $[+0.7, +4.0]$ & 0.006 \\
LLaDA-1.5 & Internal & HumanEval & Accuracy & $164$ & $13$--$0$ & $+7.9$ & $[+4.3, +12.2]$ & 0.00024 \\
LLaDA-1.5 & Internal & MBPP & Accuracy & $482$ & $14$--$10$ & $+0.8$ & $[-1.0, +2.9]$ & 0.541 \\
LLaDA-1.5 & External PRM & MATH-500 & Accuracy & $500$ & $26$--$20$ & $+1.2$ & $[-1.4, +3.8]$ & 0.461 \\
LLaDA-1.5 & External PRM & GSM8K & Accuracy & $1319$ & $44$--$39$ & $+0.4$ & $[-1.0, +1.7]$ & 0.661 \\
Dream-7B & Internal & MATH-500 & Accuracy & $500$ & $52$--$27$ & $+5.0$ & $[+1.4, +8.4]$ & 0.007 \\
Dream-7B & Internal & GSM8K & Accuracy & $1319$ & $157$--$45$ & $+8.5$ & $[+6.4, +10.5]$ & $<$0.0001 \\
Dream-7B & Internal & HumanEval & Accuracy & $164$ & $35$--$5$ & $+18.3$ & $[+11.6, +25.6]$ & $<$0.0001 \\
Dream-7B & Internal & MBPP & Accuracy & $482$ & $28$--$17$ & $+2.3$ & $[-0.4, +5.0]$ & 0.135 \\
Dream-7B & External PRM & MATH-500 & Accuracy & $500$ & $23$--$28$ & $-1.0$ & $[-3.8, +1.8]$ & 0.576 \\
Dream-7B & External PRM & GSM8K & Accuracy & $1319$ & $40$--$28$ & $+0.9$ & $[-0.3, +2.1]$ & 0.182 \\
\midrule
LLaDA-1.5 & Internal & MATH-500 & Coverage & $500$ & $61$--$14$ & $+9.4$ & $[+6.2, +12.8]$ & $<$0.0001 \\
LLaDA-1.5 & Internal & GSM8K & Coverage & $1319$ & $60$--$15$ & $+3.4$ & $[+2.1, +4.8]$ & $<$0.0001 \\
LLaDA-1.5 & Internal & HumanEval & Coverage & $164$ & $17$--$1$ & $+9.8$ & $[+4.9, +14.6]$ & 0.00014 \\
LLaDA-1.5 & Internal & MBPP & Coverage & $482$ & $23$--$8$ & $+3.1$ & $[+0.8, +5.4]$ & 0.011 \\
LLaDA-1.5 & External PRM & MATH-500 & Coverage & $500$ & $46$--$21$ & $+5.0$ & $[+1.8, +8.2]$ & 0.003 \\
LLaDA-1.5 & External PRM & GSM8K & Coverage & $1319$ & $30$--$6$ & $+1.8$ & $[+1.0, +2.7]$ & $<$0.0001 \\
Dream-7B & Internal & MATH-500 & Coverage & $500$ & $89$--$16$ & $+14.6$ & $[+10.8, +18.4]$ & $<$0.0001 \\
Dream-7B & Internal & GSM8K & Coverage & $1319$ & $178$--$28$ & $+11.4$ & $[+9.3, +13.3]$ & $<$0.0001 \\
Dream-7B & Internal & HumanEval & Coverage & $164$ & $43$--$1$ & $+25.6$ & $[+18.9, +32.3]$ & $<$0.0001 \\
Dream-7B & Internal & MBPP & Coverage & $482$ & $45$--$4$ & $+8.5$ & $[+5.8, +11.4]$ & $<$0.0001 \\
Dream-7B & External PRM & MATH-500 & Coverage & $500$ & $38$--$23$ & $+3.0$ & $[+0.0, +6.0]$ & 0.072 \\
Dream-7B & External PRM & GSM8K & Coverage & $1319$ & $48$--$13$ & $+2.7$ & $[+1.5, +3.8]$ & $<$0.0001 \\
\bottomrule
\end{tabular}
\end{center}
\end{table}

\clearpage
\section{Limitations}
\label{app:limitations}

\fixnew{All experiments sit at one operating point: two base models, $N = 16$ particles and
$L = 256$. A sweep over the particle count would mean re-decoding every benchmark under every
configuration, and at the measured cost of roughly two minutes per problem per configuration
(Appendix~\ref{app:protocol}) that grid was beyond the compute we could allocate to the full test
sets. We therefore report the operating point we can support in full, and note that the mechanism
is not specific to it: the expected fractional lineage loss under multinomial resampling is
$1 - (1 - 1/N)^{N}$ for any $N \ge 4$ (Proposition~\ref{prop:essblind}) and the floor
$\kappa = \ln N$ is defined relative to $N$, so both components are stated for general $N$. Whether
the measured margin holds at other $N$ is untested.

Task-appropriate external process-reward guidance is evaluated only on mathematics, because no
step-level code reward model met our criteria (Appendix~\ref{app:coderm}); the deliberately
out-of-distribution test of Table~\ref{tab:oodprm} is a substitute for it, not an equivalent.

The floor explicitly changes the weighted measure, so LP-SMC is unbiased for the floored target and
not for the original unfloored one, and at a boundary the floor can move at most $(N-1)/(2N-1)$ of
the normalized mass, $0.48$ at $N = 16$; the measured displacement is about an order of magnitude
smaller (Appendix~\ref{app:floor}).

Finally, coverage gains do not always become accuracy gains. The conversion depends on the
selection rule and is least reliable for the consensus rules on code, where merging by answer
equivalence has less to work with than on mathematics
(Tables~\ref{tab:selectors} and~\ref{tab:dreamselfull}).}

\end{document}